\def\year{2021}\relax
\documentclass[letterpaper]{article} % DO NOT CHANGE THIS
\usepackage{aaai21}  % DO NOT CHANGE THIS
\usepackage{times}  % DO NOT CHANGE THIS
\usepackage{helvet} % DO NOT CHANGE THIS
\usepackage{courier}  % DO NOT CHANGE THIS
\usepackage[hyphens]{url}  % DO NOT CHANGE THIS
\usepackage{graphicx} % DO NOT CHANGE THIS
\usepackage{natbib}  % DO NOT CHANGE THIS AND DO NOT ADD ANY OPTIONS TO IT
\usepackage{caption} % DO NOT CHANGE THIS AND DO NOT ADD ANY OPTIONS TO IT
\usepackage{amsmath}
\usepackage{mathtools}
\usepackage{amssymb}
\usepackage{algorithm} 
\usepackage{algorithmic} 

\graphicspath{{img/}}
\title{Meta-Reinforcement Learning for Heuristic Planning}
\author{Ricardo Luna Gutierrez, Matteo Leonetti \\ 
}
\affiliations {
	University of Leeds\\
	\{scrlg, m.leonetti\}@leeds.ac.uk
}
\begin{document}

\maketitle

\begin{abstract}
	Heuristic planning has a central role in classical planning applications and competitions. Thanks to this success, there has been an increasing interest in using Deep Learning to create high-quality heuristics in a supervised fashion, learning from optimal solutions of previously solved planning problems. Meta-Reinforcement learning is a fast growing research area concerned with learning, from many tasks, behaviours that can quickly generalize to new tasks from the same distribution of the training ones. We make a connection between meta-reinforcement learning and heuristic planning, showing that heuristic functions meta-learned from planning problems, in a given domain, can outperform both popular domain-independent heuristics, and heuristics learned by supervised learning. Furthermore, while most supervised learning algorithms rely on ad-hoc encodings of the state representation, our method uses as input a general PDDL 3.1 description. We evaluated our heuristic with an A* planner on six domains from the International Planning Competition and the FF Domain Collection, showing that the meta-learned heuristic leads to the expansion, on average, of fewer states than three popular heuristics used by the FastDownward planner, and a supervised-learned heuristic. 
\end{abstract}

\section{Introduction}

Developing planning heuristics that allow forward search algorithms to generate high quality plans is a constant focus in AI planning research. This interest is justified by the success of heuristic search algorithms in applications~\cite{Edelkamp2012} and at the International Planning Competition~\cite{Hoffmann201,Helmert2006,Franco2018}. A central component of these planning algorithms is the heuristic function, which estimates the cost of reaching the goal from any state. This estimate guides the planner to low cost states discarding unpromising regions of the state space. 

While some of the most successful heuristic search approaches are domain-independent~\cite{Hoffmann201,Helmert2006}, there has been a raising interest in making use of machine learning for domain-dependant heuristics \cite{Yoon2008,Garrett2016,Gomoluch2017,Groshev2018,Toyer2018,Shen2020}. These approaches are based on supervised learning, and learn from the optimal plans of previously solved planning problems. A crucial problem of heuristic learning is generalizing across different instances of the same planning domain, so that previous instances can inform the search on new, unseen, instances. Learning from optimal plans implies obtaining only information about a very limited area of the state space (the states along the optimal plan), and requires a high number of solved planning instances to achieve satisfactory generalization.

Reinforcement Learning (RL) algorithms, on the other hand, learn a \emph{value function} for tasks modelled as Markov Decision Processes (MDPs). The value function guides the exploration of the agent just like a heuristic function for the planner, but the estimate of the value function is made increasingly accurate through learning, and eventually converges to the lowest cost (or equivalently, highest reward) from any state to a goal state. Since the optimal value function is also the best possible heuristic, it seems natural to consider reinforcement learning as a method to learn heuristic functions. Value functions, however, are specific to a particular task, and a considerable amount of research in RL is devoted to achieving generalization across tasks, so that the knowledge gathered in a task can be reused in a new, similar, task. Meta-reinforcement learning~\cite{Vanschoren2018} is a recent and fast growing field tackling this challenge. A set of tasks is used to train the agent so that common features of the tasks may be learned and leveraged to bootstrap learning on new, unseen, tasks.

We make the first connection, to the best of our knowledge, between meta-reinforcement learning (meta-RL) and heuristic classical planning. We propose to learn heuristic functions for a given planning domain through meta-reinforcement learning, and show that the learned value function generalizes effectively in a collection of domains from the International Planning Competition (IPC) and the FF Domain Collection. We evaluated our heuristic using A*~\cite{Hartet1968}. Our results show that the meta-RL heuristic can outperform popular heuristics such as $h^{max}$, $h^{add}$ and LM-cut \cite{helmert2009}, and a supervised learning approach, by expanding a smaller number of nodes at search time.

\section{Related Work}
\label{sec:related}

Research on meta-RL has been quickly growing in the past few years. Applications such as navigation tasks \cite{Finn2017,Duan2016,Wang2017,Mishra2018}, classic control tasks \cite{Li2018,Schweighofer2003,Xu2018,Sung2017}, and locomotion tasks \cite{Finn2017,Rakelly2019,gupta2018meta}, have shown the potential of meta-reinforcement learning to quickly adapt a policy to a new task generated from the same distribution as the training tasks. To the best of our knowledge, meta-RL has never been used to learn heuristics for planning.

There has also been an increasing interest in developing deep learning techniques to improve the performance of automated planning \cite{Fern2011}. For instance, learning policies \cite{Garg2020,Groshev2018,Toyer2018,Issakkimuthu2018,Garg2019,Garg2020,Shen2019GuidingSW}, planner selection \cite{Sievers2019,Ma2020,Katz2018} and heuristics \cite{Arfaee2011,Groshev2018,Samadi2008,Thayer2011,Garrett2016,Shen2020} have been widely explored. Our work fits within the heuristic learning category.

Recent methods for learning heuristics combine or improve on existing heuristics \cite{Arfaee2011,Groshev2018,Samadi2008,Thayer2011,Garrett2016,Shen2020}.  All of these methods use supervised learning but differ in the encoding of the states, proposing, for instance, the use of images \cite{Groshev2018,Ma2020,Katz2018} or sophisticated network models \cite{Shen2020,Toyer2018}. A common approach is to do regression on the heuristic values obtained from pre-computed plans~\cite{Shen2020,Toyer2018,Garrett2016,Yoon2008}, and for this reason, it is the baseline we used to compare against supervised methods.

In this work, we compare meta-RL against a supervised learning approach without the use of any domain-specific encoding. Instead, we train the network by using a numeric representation taken from a domain description in PDDL 3.1. Specific domain-dependent representations may yield even better results. We propose to use meta-RL instead of supervised learning, and to learn by exploration on a limited number of instances from the planning domain, rather than from precomputed plans. The exploration allows the agent to learn values of sub-optimal states as well, which we hypothesise to provide valuable information to transfer.

\section{Background and Notation}

We consider the classical RL setting, where a task is represented as a Markov Decision Process (MDP) $m= \langle S, A, R, P, \gamma, \mu \rangle$, where $S$ is the set of states, $A$ is the set of actions, $R \subseteq \mathbb{R}$ is the set of possible rewards, $P(s', r | s, a)$ is a joint probability distribution over next state and reward given the current state and action, $0 \leq \gamma \leq 1$ is the discount factor, and $\mu(s)$ is the initial state distribution. We assume that tasks are \emph{episodic}, that is, the agent eventually reaches an \emph{absorbing} state that can never be left, and from which the agent only obtains rewards of $0$. 

The behaviour of the agent is represented by a policy $\pi(a|s)$ returning the probability of taking action $a$ in state $s$. 
  
The goal of the agent is to compute an optimal policy $\pi^*$, which maximizes the expected return $ \mathbb{E}_\pi [ G_t ] = \mathbb{E}_\pi [ \sum_{i \geq t} \gamma^{i-t} R_{i+1}]$ from any state $S_t$ at time $t$. The value function $v_{\pi}(s) = \mathbb{E}_{\pi} [G_t]$ represents the expected return obtained by choosing actions according to policy $\pi$ starting from state $S_t = s$. The objective of minimizing cost, rather than maximizing reward, can be represented simply by having negative rewards. An actor-critic architecture explicitly represents with function approximators both the policy (actor) and the value function (critic). The policy is trained so as to maximize its value, by gradient ascent, while the value is trained to minimize its prediction error, by gradient descent. 

\subsection{Meta-Reinforcement Learning}
\label{sec:meta-learning}
In the Meta-Reinforcement Learning framework, an agent has access to series of different tasks to train on, enabling it to gain broad information about the domain and adapt quickly to new tasks. Given a potentially infinite set tasks $\mathcal{M}$ and a distribution over the tasks $p(\mathcal{M})$, the agent is presented $T$ tasks $\mathcal{T} = \{m_i\}_{i=1}^{T}$ to train on. At test time, a new task $m_j \sim p(\mathcal{M})$ is extracted from $p(\mathcal{M})$. This task has not been seen by the agent before, $m_j \notin \mathcal{T}$, and the agent is expected to adapt \emph{quickly} and achieve good performance in this new task. 

The commonly assumed meta-RL framework described above does not guarantee that the set of tasks in $\mathcal{M}$ are at all related, and that transfer is even possible. This is in general left to the intuition of the designer, and much ingenuity has been used in existing meta-RL applications to create appropriate set of tasks and corresponding distributions $p(\mathcal{M})$.  We use PDDL domain descriptions to specify the set of tasks, which therefore all belong to the same planning domain, and PDDL problem descriptions to define single training and test tasks. It is possible that the PDDL description can be optimized for learning, but in this work we used the PDDL descriptions of all domain as they have been used in either IPC, or the FF Domain Collection, without any specific tweaking.

Meta-RL is usually divided into two categories \cite{Rakelly2019}: gradient-based methods and context-based methods. Gradient-based approaches learn from sampled transitions from the tasks using hyperparameters, meta-learned loss functions or policy gradients \cite{Finn2017,Xu2018,Xu22018,Houthooft2018,Sung2017,Stadie2018}. In context-based approaches, on the other hand, models are trained to use previous states and actions as a form of task-specific \emph{context} \cite{Duan2016,Wang2017,Mishra2018}. In gradient-based methods the agent adapts to the new tasks through further online learning, while context-based methods only require to fill a memory buffer with transitions from the new task, which are used to identify the new context, and no further learning is required. 

Context-based methods are more suitable for planning heuristics, since they do not require any exploration or learning during deployment. We chose RL$^2$ \cite{Duan2016,Wang2017} as a suitable context-based method, using memory in a recurrent neural network. RL$^2$ does not update its parameters at test time, but instead integrates experience through its memory.

\subsection{RL$^2$}
\label{sec:RL2}

In the RL$^2$ framework the previous reward $r_{t-1}$ and previous action $a_{t-1}$ are integrated with the current state $s_{t}$ to form the observation, at time step $t$, to be fed to the training model. The purpose is to allow the model to learn the interactions between sates, actions and rewards in the current domain, and build a context that identifies key properties of the task at hand. $RL^2$ makes use of an LSTM in which its hidden states serve as memory for monitoring the observed trajectories. The agent is trained with a set of different tasks in an episodic manner. At the start of each training episode, a task $m \in \mathcal{T}$ is sampled and the internal state of the LSTM is reset. The agent then interacts with the environment executing its policy and collecting experience (in the form of state, action, and reward samples) as results of these interactions. The collected experience is used to train the network weights to learn a policy and a value function that enable the agent to maximize the sum of the rewards obtained over all the episodes. At test time, adjustment of the value function to the new task takes place observing the first few transitions, as the internal memory fills up and creates the context for the estimate of the value function in current the task.

The learning method used to improve the policy and the value function is not specified in the $RL^2$ framework. We used the popular Proximal Policy Optimization (PPO) \cite{Schulman2017} method as a learning algorithm, with an actor-critic architecture. PPO is a learning algorithm that optimizes the policy $\pi_\theta$, represented as a neural network with parameters $\theta$, using gradient ascent on the objective function: 
\begin{multline*}
L(s_t,a_t,\theta_k,\theta)=\mathbb{E}[\frac{\pi_\theta(a_t|s_t)}{\pi_{\theta_k}(a_t|s_t)}A^{\pi_{\theta_k}}(a_t|s_t)), \\ clip(\frac{\pi_\theta(a_t|s_t)}{\pi_{\theta_k}(a_t|s_t)},1-\epsilon,1+\epsilon)A^{\pi_{\theta_k}}(a_t|s_t)],
\end{multline*}
where $\pi_{\theta_k}$ is the policy at the start of an episode, before the weights are updated, and $A$ is the estimated advantage given by:
\begin{equation*}
A^{\pi_{\theta_k}}(a_t|s_t) = r_{t+1} + v_{\pi_{\theta_k}}(s_{t+1}) - v_{\pi_{\theta_k}}(s_{t}).
\end{equation*}
The hyperparameter $\epsilon$ ensures that the policy does not change drastically, and takes small values, usually in $[0.1, 0.2]$. Two neural networks are used to represent the critic and the actor.

\subsection{Task Selection}

In the standard meta-RL training approach dense sampling of the task set is used to generate a large set of training tasks. Although dense sampling has shown good results \cite{Duan2016,Wang2017,Finn2017} it may be difficult or too expensive to generate hundreds of tasks for training. While more tasks are believed to be beneficial in the limit, task selection has been recently shown to be effective~\cite{Luna2020} when dense sampling is not feasible. We employed Information-Theoretic Task Selection (ITTS)~\cite{Luna2020} to limit the number of tasks on which to meta-train, and make the most out of the generated tasks. ITTS filters a set of training tasks $\mathcal{T}$ by selecting tasks that are different between each other and relevant to a subset of tasks sampled from the task distribution, which they refer to as \emph{validation} tasks $\mathcal{F}$. The result is a smaller training set $\mathcal{C}\subseteq \mathcal{T}$ that achieves better performance than the original set $\mathcal{T}$. 

In ITTS  the difference between two tasks $m_1$ and $m_2$ is calculated as the average KL divergence of their optimal policies over states sampled from the validation tasks:

\begin{multline*}
\delta(m_1, m_2) \coloneqq \frac{1}{\lvert \mathcal{F} \rvert} \sum_{m_j \in \mathcal{F}}  \frac{1}{\lvert S_j \rvert} \sum_{s \in S_j} \mathcal{D}_{KL} \left( \pi^*_{m_1}(a|s) \right. \parallel \\
\left. \pi^*_{m_2}(a|s) \right).
\end{multline*}

To calculate the relevance of $m_1$ to a validation task $m_2$, in ITTS the optimal policy of $m_1$, $\pi_{m_1}^*$ is transferred to $m_2$, and the agent trained on $m_2$ for $l$ episodes. The relevance is defined as the expected difference in entropy of the two policies before and after training over the states of the validation tasks:
\begin{multline*}
\label{eq:relevance}
\rho_l(m_1, m_2) \coloneqq  \mathbb{E}_{s \sim d^{m_2}_{\pi^*_{m_1}} , \pi_{m_1,m_2}^l} \left[ H\left(\pi_{m_1}^*(a|s) ) \right. \right. \\ 
- \left. \left. H(\pi_{m_1,m_2}^l(a|s)\right)\right],
\end{multline*}
where $\pi_{m_1,m_2}^l$ is the policy obtained from training in $m_2$ starting from $\pi_{m_1}^*$ after $l$ episodes, and $d^{m_2}_{\pi^*_{m_1}}$ is the on-policy distribution of the optimal policy of task $m_1$ when applied to task $m_2$. The on-policy distribution of a policy in an MDP is the probability of visiting each state of the MDP while following the given policy. Intuitively, the relevance measure determines whether starting learning from the optimal policy of a task $m_1$, and learning for a number of episodes $l$ in a task $m_2$, results in a policy of lower entropy (and therefore an information gain) than the initial policy. Tasks $m_i$ are added to the set of selected tasks $\mathcal{C}$ if they are (1) different from all the other tasks currently in $\mathcal{C}$ by at least a (domain-dependent) parameter $\epsilon$, and (2) relevant to at least one validation task $v_j$ so that $\rho_l(m_i, v_j) \geq 0$.

\subsection{Planning}

In this work, we consider learning heuristics for classical planning, that is, for deterministic, sequential planning problems. The classical MDP-based RL framework is more general, and permits immediate extensions to probabilistic planning. One of the most popular ways to represent deterministic and fully observable planning tasks is PDDL (Fox and Long 2003;Helmert 2009). The domains we used are based on a subset of PDDL 3.1.

A planning task defined in PDDL is divided in two parts: domain description and problem description. A planning domain can be described as a tuple $\langle P,A \rangle$ where $P$ is a set of predicates and $A$ is a set of parametrized actions. These actions are constrained by a set of precondition predicates that must be satisfied in order for the action to be executed, and contain a description of the effects that will be applied to the current state in the case of a successful execution. A planning problem is defined as $\langle O,I,G,c \rangle$ where $O$ is a set of objects in the domain, $I$ is the initial state, $G$ is a set of goals, and $c(s,a,s')$ is the cost of the transition landing in $s'$ after taking action $a$ in state $s$. For shortest planning problems we consider $c(s,a,s')=1$ for every transition. The initial state describes the combination of objects $O$ and predicates $P$ that are true before any action has been taken.

A heuristic function $h(s)$ estimates the cost of reaching the goal from state $s$, and is used to drive the planning search by choosing states with low cost estimations. A heuristic is considered admissible when it never overestimates the cost of achieving a goal. Since the optimal value function returns the expected cost to reach the goal under the optimal policy, $h(s) = v^*(s)$ would be a perfect heuristic, and A* would only expand states along the optimal plan by using it.

\section{Learning Planinng Heuristics}

We take advantage of the generalization properties of an existing meta-RL method, RL$^2$, and apply it to classical planning, with the aim of learning a heuristic function that gets as close as possible to $v^*$. The methodology is summarized in algorithm \ref{alg:hmrl} and consists in the following steps.

\subsection{Learning Problem Definition}

A numerical representation of the state space and the reward function must be defined, so that they can be used in the input and objective function of the neural networks encoding the policy and the value function. We define a numerical vector $s_i = \langle s_i^{(1)}, s_i^{(2)}, \ldots, s_i^{(n)} \rangle$, representing the state at time step $i$, from a domain description in PDDL 3.1. This vector is created by concatenating boolean predicates and numeric variables that compose the problem, such as object positions, goal positions, binary state of an object, goals achieved, and so on. Each predicate is translated into a value in  $\{0,1\}$, while numeric variables are directly added to the state representation.

The cost function of the planning problem maps onto the reward function of the learning problem, so that $P(s',r| s,a) =1 $ if $r = -c(s,a,s')$, and $P(s',r| s,a) =0$ otherwise. For shortest planning problems, we used a reward of $-1$ per action. We also used a positive reward $r_G$ for reaching a goal state, to distinguish this condition from other possible ways to end an episode (for instance, the agent reaching a dead end, or maximum number of actions executed).

\begin{algorithm}[ht]
	\caption{Meta-Reinforcement Learning for Heuristic Planning}
	\label{alg:hmrl}
	\begin{algorithmic}[1]
		\STATE {\bfseries Input:} $\mathcal{N}$ number of tasks to generate
		
		\STATE {\bfseries Output:} Meta-trained value function $v_\theta$

		\STATE $\mathcal{T} \leftarrow $ Generate $\mathcal{N}$ tasks
		\STATE Learn the optimal policies for all the tasks in $\mathcal{T}$ using PPO
		
		\STATE Run ITTS on $\mathcal{T}$ to obtain a subset of optimal tasks $\mathcal{C}$
		
		\STATE Meta train on parameters $\theta$ with $RL^2$ using tasks $\mathcal{C}$ to obtain $v_\theta$. 		
		
	\end{algorithmic}
\end{algorithm}

\subsection{Training Task Generation and Selection}

The first step in the methodology consists in generating a number of training task candidates, forming the initial set of tasks $\mathcal{T}$. This step is in general domain dependent. A common way relies on a parametrized description of the domain, so that a distribution can be defined over the parameters' range. Examples of this method for task generation are discussed in Section Experimental Evaluation.

The set of candidate tasks can be used entirely, or filtered to identify a subset that leads to better knowledge transfer. ITTS can be used at this stage to select a subset $\mathcal{C}$ of the available tasks $\mathcal{T}$. This step requires learning the optimal policy of all the tasks, which is the main computational bottleneck. ITTS is not essential for the methodology, but it did further improve our results, as has been shown for other meta-RL tasks where dense sampling is not possible or desirable~\cite{Luna2020}.

\subsection{Model Training}

The meta-RL model is trained using $\mathcal{C}$ as a training set and Proximal Policy Optimization (PPO) \cite{Schulman2017} as the learning algorithm, as described in Section RL$^2$.

The meta-trained value function $v_\theta$, depending on a parameter vector $\theta$, is an adaptive estimate of the value $v^*$ for every task in the domain. The adaptation takes place over the initial transitions, when the memory of the LSTM network fills up, providing the context of the new task. When learning converges on the training tasks, the learned value function can be used to define the planning heuristic, so that $h^{MRL}(s) = -v_{\theta}(s)$. The optimal value function for a given task is an admissible heuristic for that task. The meta-learned heuristic, however, is subject to function approximation and generalization across tasks, therefore it may not be admissible.

\section{Experimental Evaluation}

The experimental evaluation aims at: (1) showing that the meta-learned heuristic leads to the expansion of fewer states than both popular domain-dependent heuristics and a supervised-learned heuristic with the same state representation; (2) evaluating the quality of the generated plans, since the heuristic is not admissible in general, and may return suboptimal plans; (3) determining how many more tasks a supervised learning approach needs to reach a comparable performance.

As baselines, we used no heuristic (Blind), $h^{max}$, $h^{add}$ and LM-cut, from FastDownward~\cite{Hoffmann201,Helmert2006}. We also  used a supervised learning approach, $h^{SUPER}$, derived from existing approaches (cf. Section Related Work), but using the same PDDL-based state representation as the meta-RL heuristic. This allows us to compare meta-RL with supervised learning without any specific representation tuning for either method.

For the supervised-learned heuristic, we followed the popular approach of using optimal plans to obtain the real heuristic $h^*(s)$ of each state $s$ of the plan. To obtain these optimal plans we used FastDownward. The network is trained using $s$ as input and $h^*(s)$ as the target. We tuned the hyperparameters and architecture of the model for each domain, and selected the best performing combination.

\subsection{Domains}

The domains used for evaluation are benchmarks of the classical planning track of IPC and the FF Domain Collection, some of which have also been used to evaluate deep learning approaches in planning. Tasks were generated by drawing their parameters uniformly at random as reported for each domain below. In addition, we also randomly sampled $5$ validation tasks for ITTS, as in the original publication~\cite{Luna2020}, and a number of test tasks. The number of training and test tasks varies for each domain, as some allow more parameter combinations than others. At least $10$ test tasks were used for every domain. The learning process was repeated $5$ times for both meta-RL and supervised learning, and all graphs show mean and standard deviation.

\subsubsection{Snake}

Snake was introduced in IPC-2018. In this domain, a snake navigates a grid with the goal of eating apples that are spawned in the grid. Each time an apple is eaten, the snake extends its length by one and a new apple spawns until there are no more apples left. The grid has a fixed size of 6x6 with $15$ apples in all instances. The grid is initialized with $5$ apples, while the others spawn as the snake eats the available ones. The position of the apples and the initial position of the head and tail of the snake are randomly selected. A total of $20$ training tasks were generated, of which ITTS selected $12$. A total of $15$ instances were used for testing.

\subsubsection{Sokoban}

Sokoban is a game from IPC-2008, also used to evaluate previous Deep Learning approaches \cite{Shen2020,Groshev2018}. The agent must push objects around a grid with the goal of moving them to specific locations. The grid has size 5x5 with $2$ objects and $3$ obstacles in all instances. The initial positions of the agent, objects, and obstacles, and the position of the goals (one per object) are randomly selected.
%The initial agent, objects and goals position were randomly changed to generate all the problems. 
A total of $20$ training tasks were generated, of which ITTS selected $11$. A total of $12$ instances were used for testing.

\subsubsection{Gripper}

Gripper is a modification of the domain used in IPC-1998, as used in previous work for learning heuristic functions \cite{Shen2020}. In this domain there is a robot with two grippers that can carry an object each. The goal is to move a certain number of objects from one room to another. All instances have $2$ rooms and $4$ objects. The initial location of the objects and the robot, as well as the target room, are randomly selected. A total of $15$ training tasks were generated, of which ITTS selected $10$. A total of $10$ instances were used for testing.

\subsubsection{Blocksworld}

Blocksworld is a popular planning domain from IPC-2001, also used to evaluate previous approaches \cite{Shen2020}. A set of blocks lie on a table. The goal is to build stacks of blocks. Only one block can be moved at a time. All instances had $4$ blocks. Each block can be free, under another block, and/or over another block. The initial configuration of the blocks as well as the goal configuration are randomly selected. $15$ training tasks were generated, of which ITTS selected $10$. A total of $10$ instances were used for testing.

\subsubsection{Ferry}

This domain was extracted from the FF Domain Collection, because it has been used to evaluate previous approaches \cite{Shen2020,Sievers2019}. In this domain, a ferry must transport a certain number of cars from their start location to a goal location. The ferry can only carry one car at a time. In all instances there are $4$ cars and $4$ locations. The initial positions of the cars and the ferry, as well as the goal position of the cars are randomly selected. A total of $15$ training tasks were generated, of which ITTS selected $12$. A total of $10$ instances were used for testing.

\subsubsection{Nurikabe}

Nurikabe was introduced in IPC-2018. In Nurikabe a robot must paint a certain pattern in a grid. The robot cannot move into locations that have been painted or have already been assigned to a non-painted block. All instances have a grid of size 5x5, $2$ colours (black or white), and each colour has $3$ or $4$ assigned cells. The initial position of the robot, of the sources (where the robot can start painting), and the position of the cells that must be painted are randomly selected. A total of $15$ training tasks were generated, of which ITTS selected $10$. A total of $10$ instances were used for testing.

\begin{figure*}
	\centering
	\begin{tabular}{ccccc}
		& \hfil \LARGE{Nodes Expanded} & \hfil \LARGE{Normalized Plan Length} \\
		\raisebox{.9\height}{\rotatebox{90}{\LARGE{Snake}}}&
		\includegraphics[width=8cm, height=3.3cm]{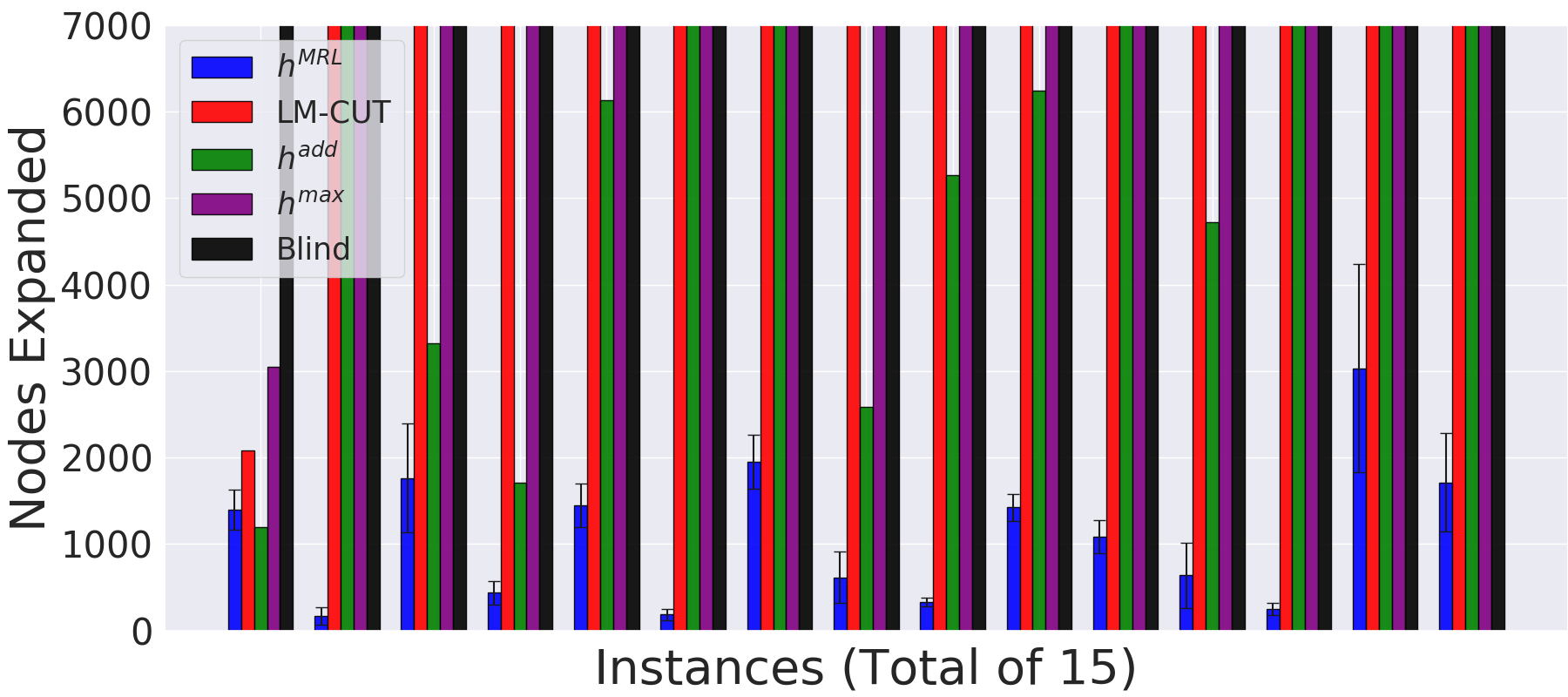}&
		\includegraphics[width=8cm, height=3.3cm]{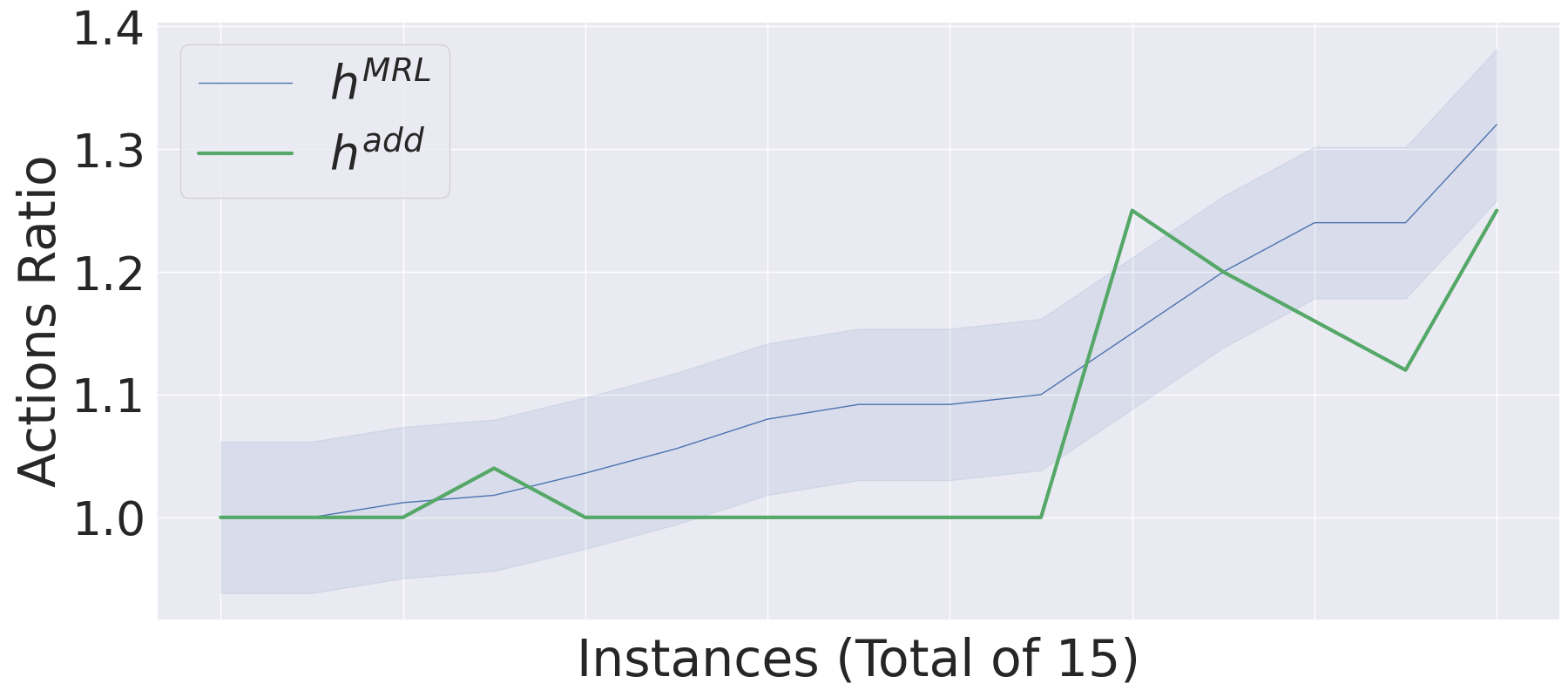}& \\
		\raisebox{.5\height}{\rotatebox{90}{\LARGE{Sokoban}}} &
		\includegraphics[width=8cm, height=3.3cm]{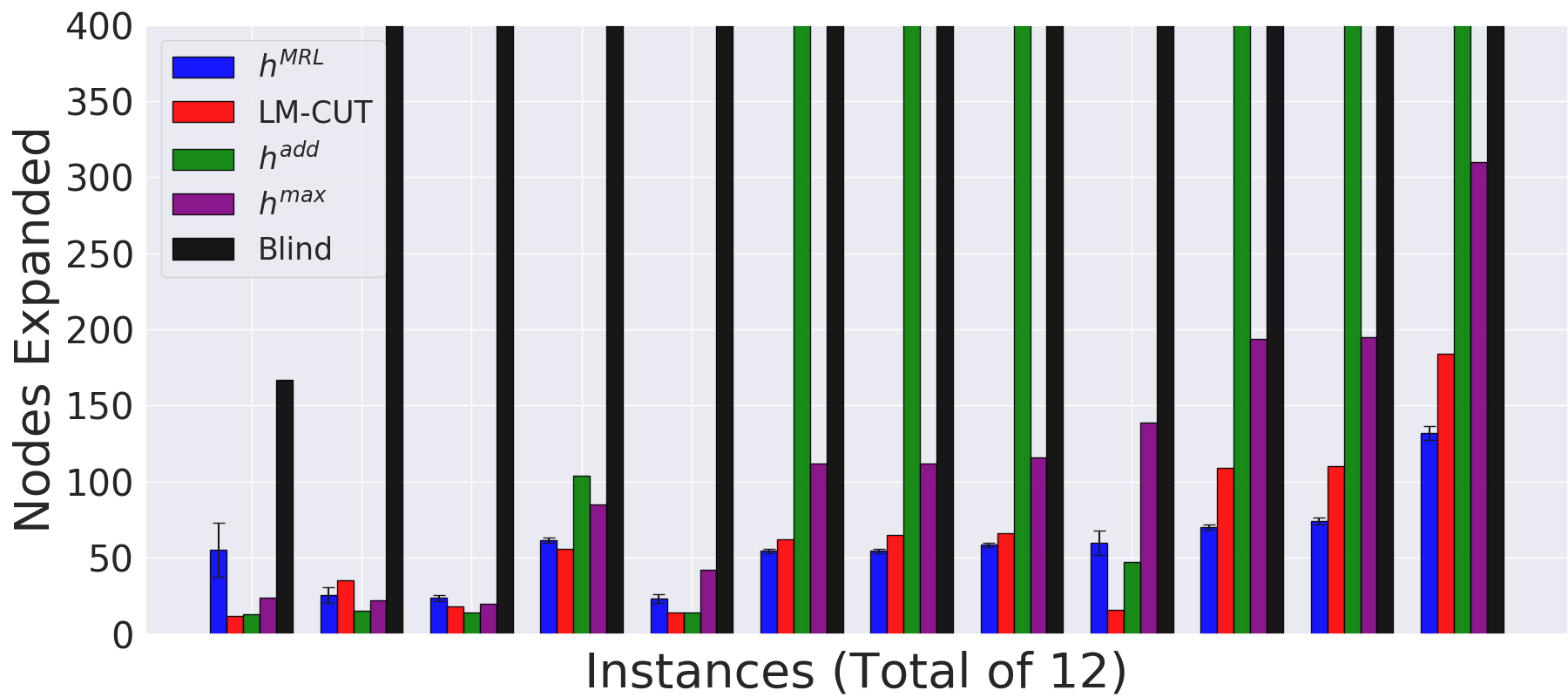}&
		\includegraphics[width=8cm, height=3.3cm]{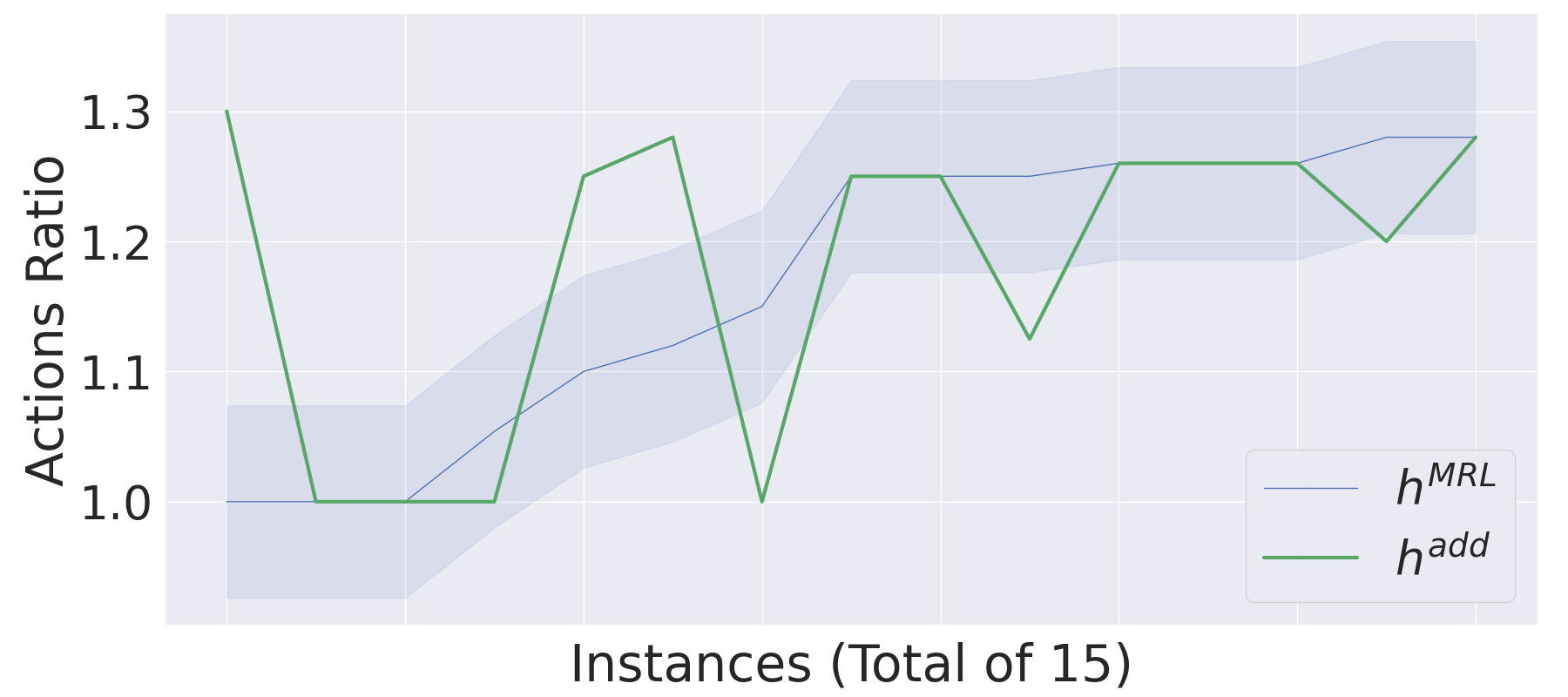}&\\
		\raisebox{.9\height}{\rotatebox{90}{\LARGE{Blocks}}} &
		\includegraphics[width=8cm, height=3.3cm]{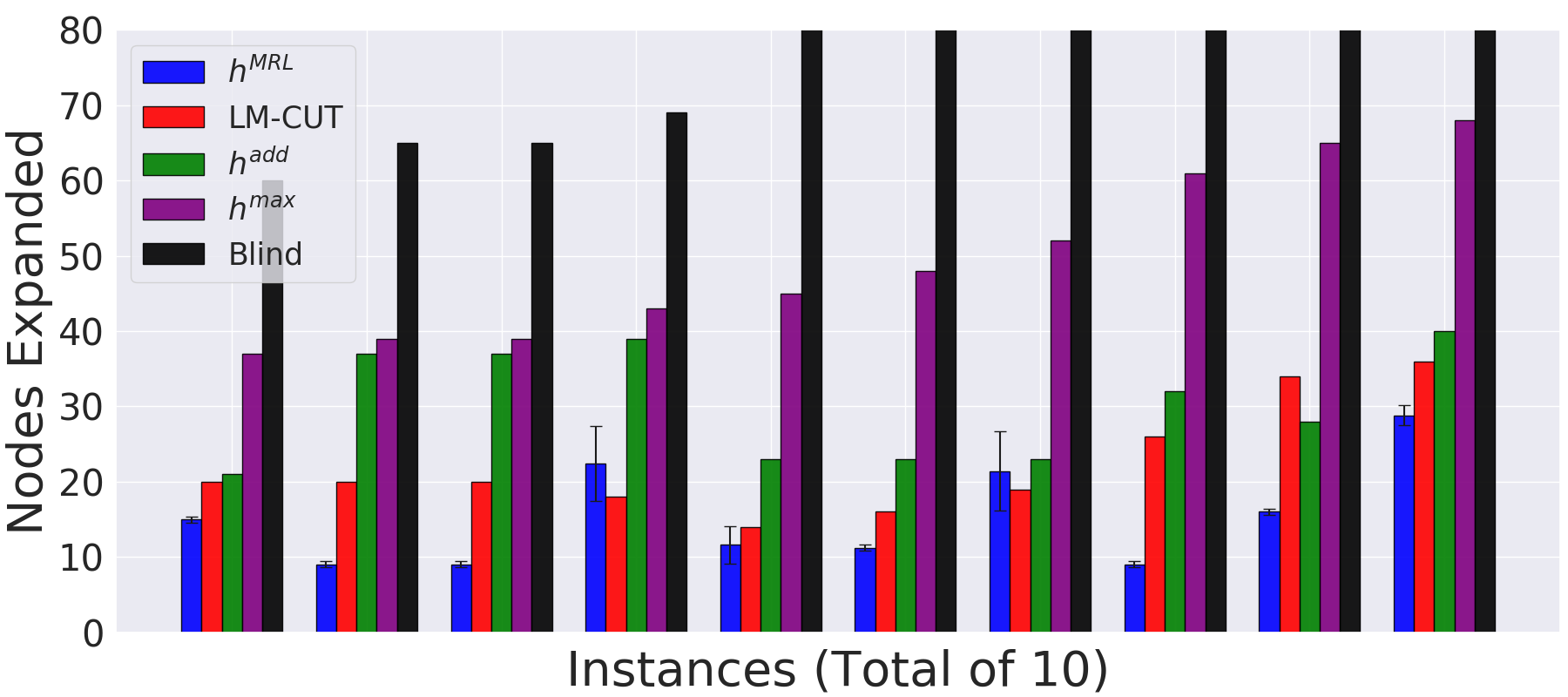}&
		\includegraphics[width=8cm, height=3.3cm]{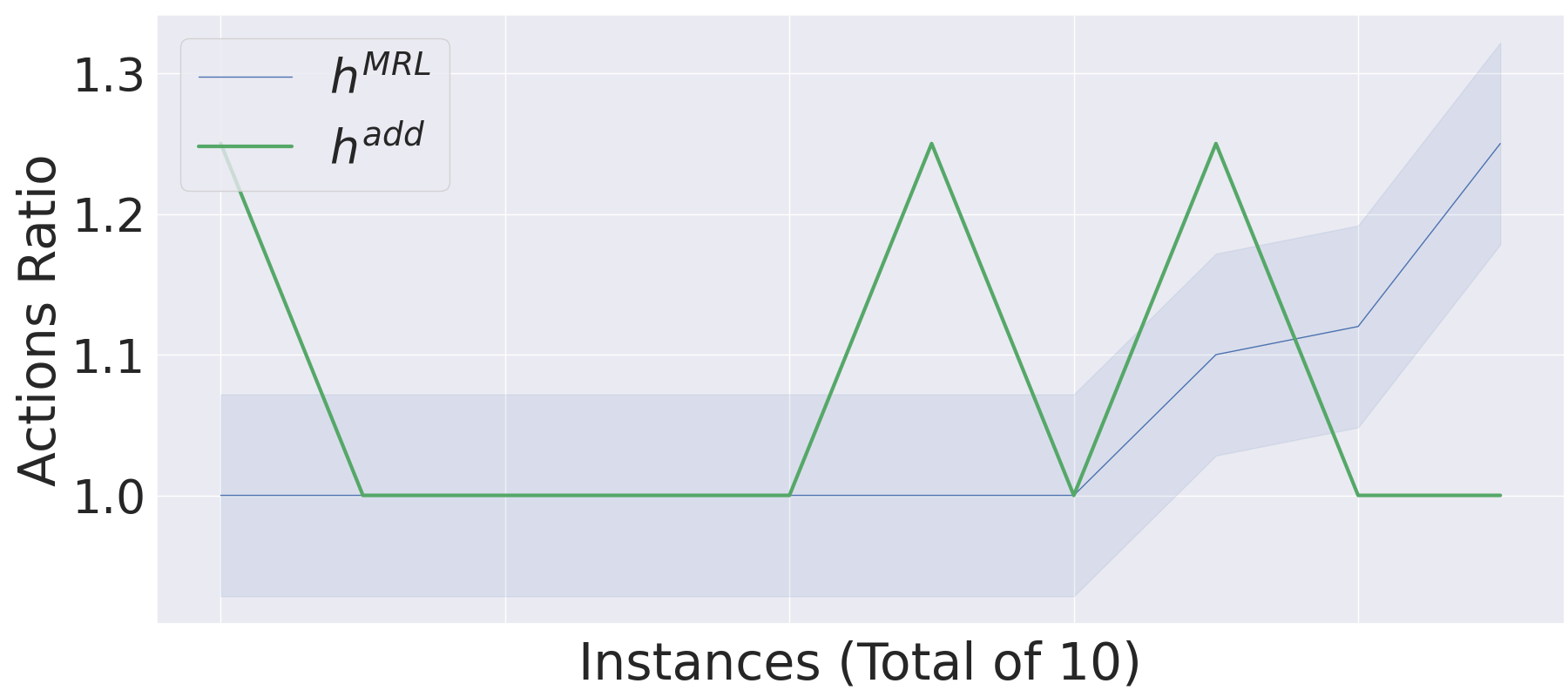}& \\
		\raisebox{.5\height}{\rotatebox{90}{\LARGE{Nurikabe}}} &
		\includegraphics[width=8cm, height=3.3cm]{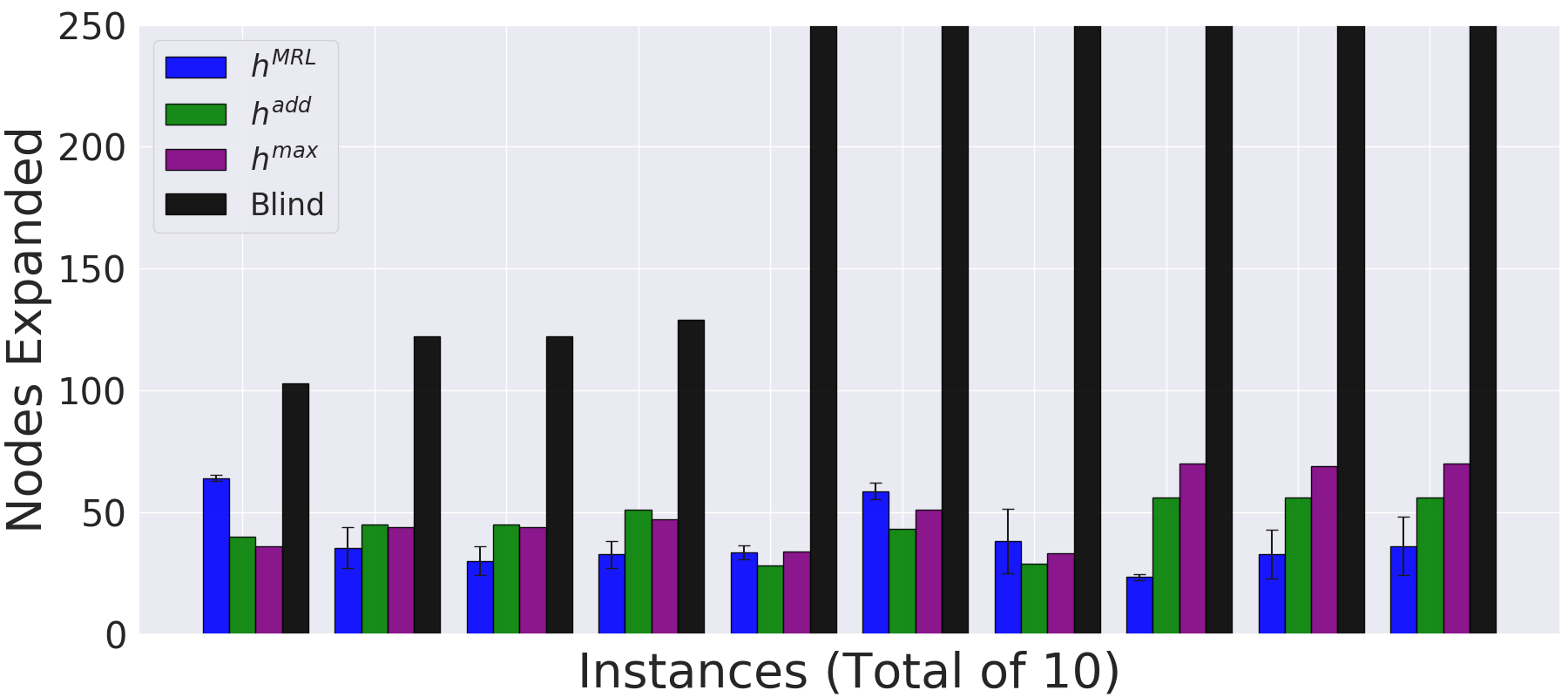}&
		\includegraphics[width=8cm, height=3.3cm]{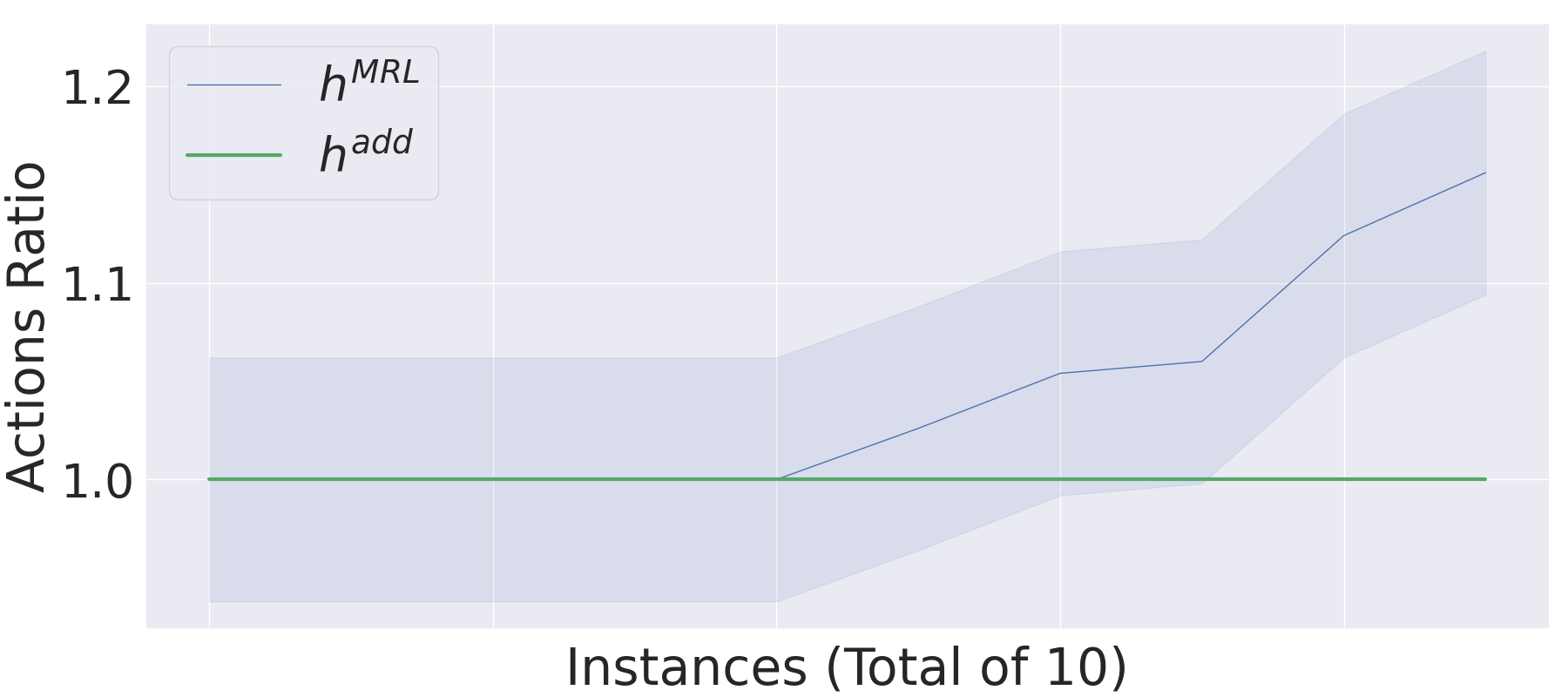}& \\
		\raisebox{1.2\height}{\rotatebox{90}{\LARGE{Ferry}}} &
		\includegraphics[width=8cm, height=3.3cm]{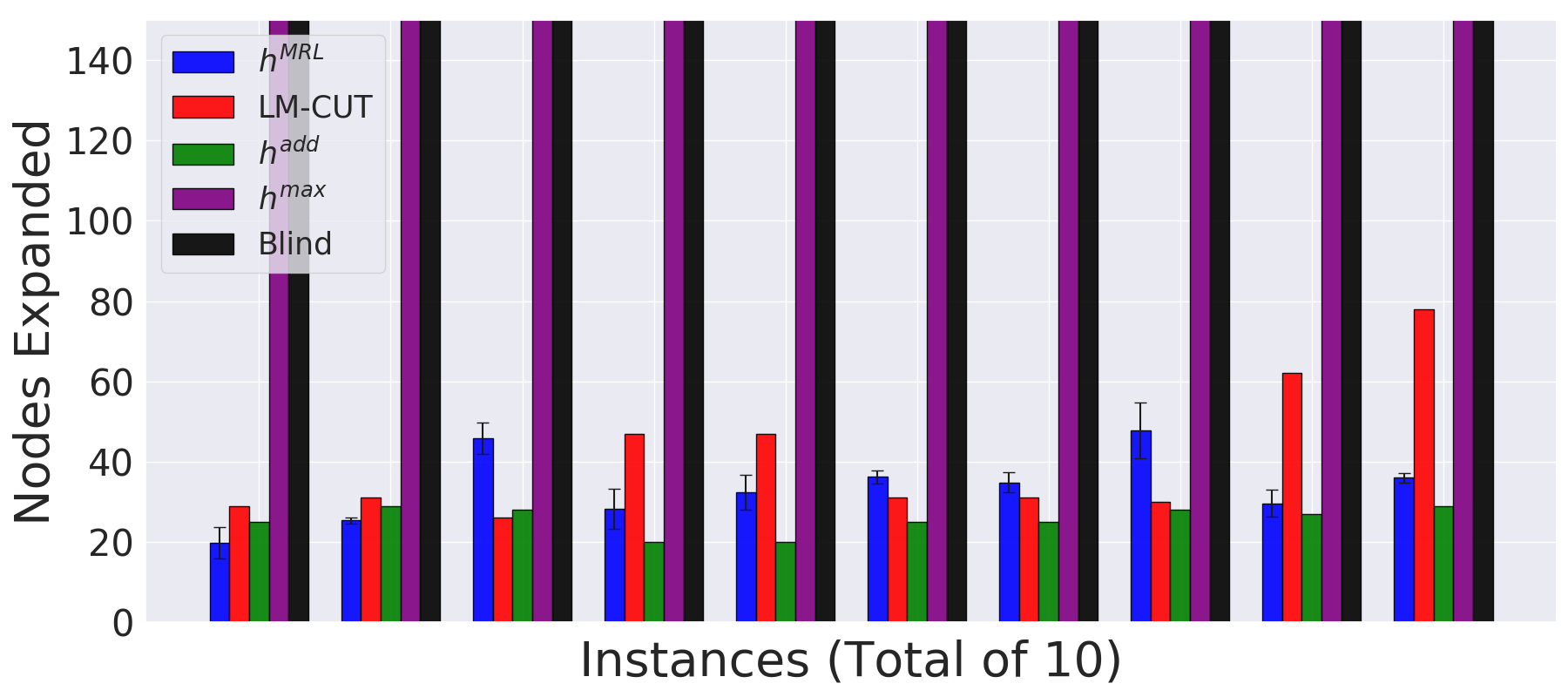}&
		\includegraphics[width=8cm, height=3.3cm]{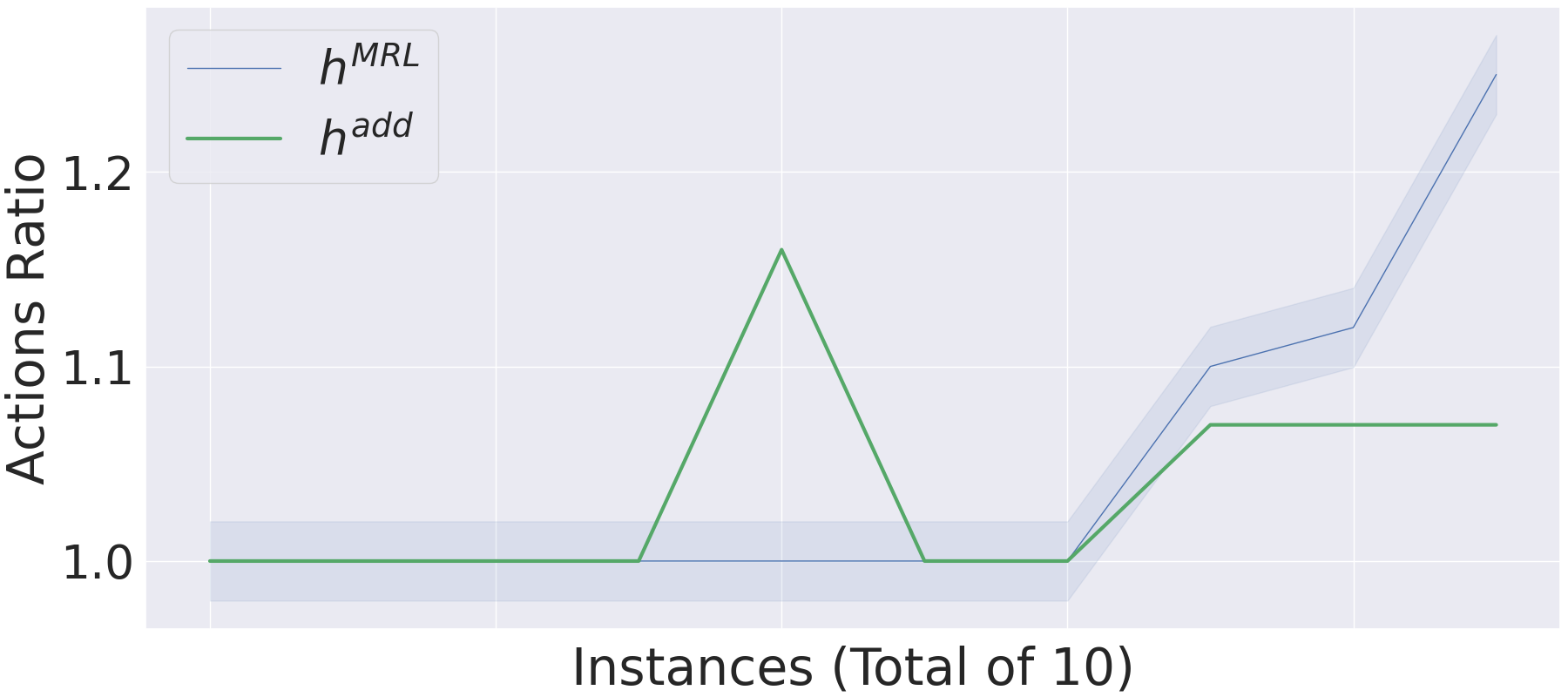}&\\
		\raisebox{.7\height}{\rotatebox{90}{\LARGE{Gripper}}} &
		\includegraphics[width=8cm, height=3.3cm]{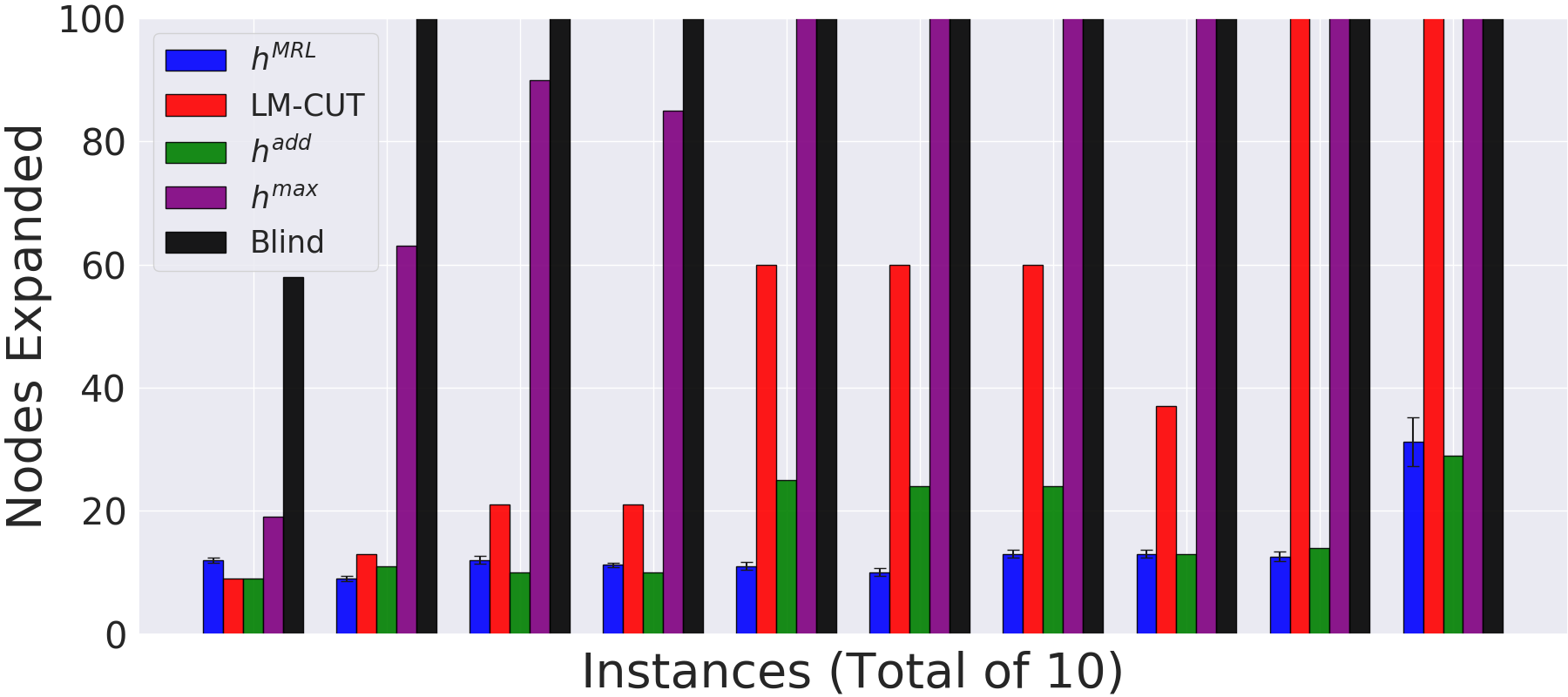}&
		\includegraphics[width=8cm, height=3.3cm]{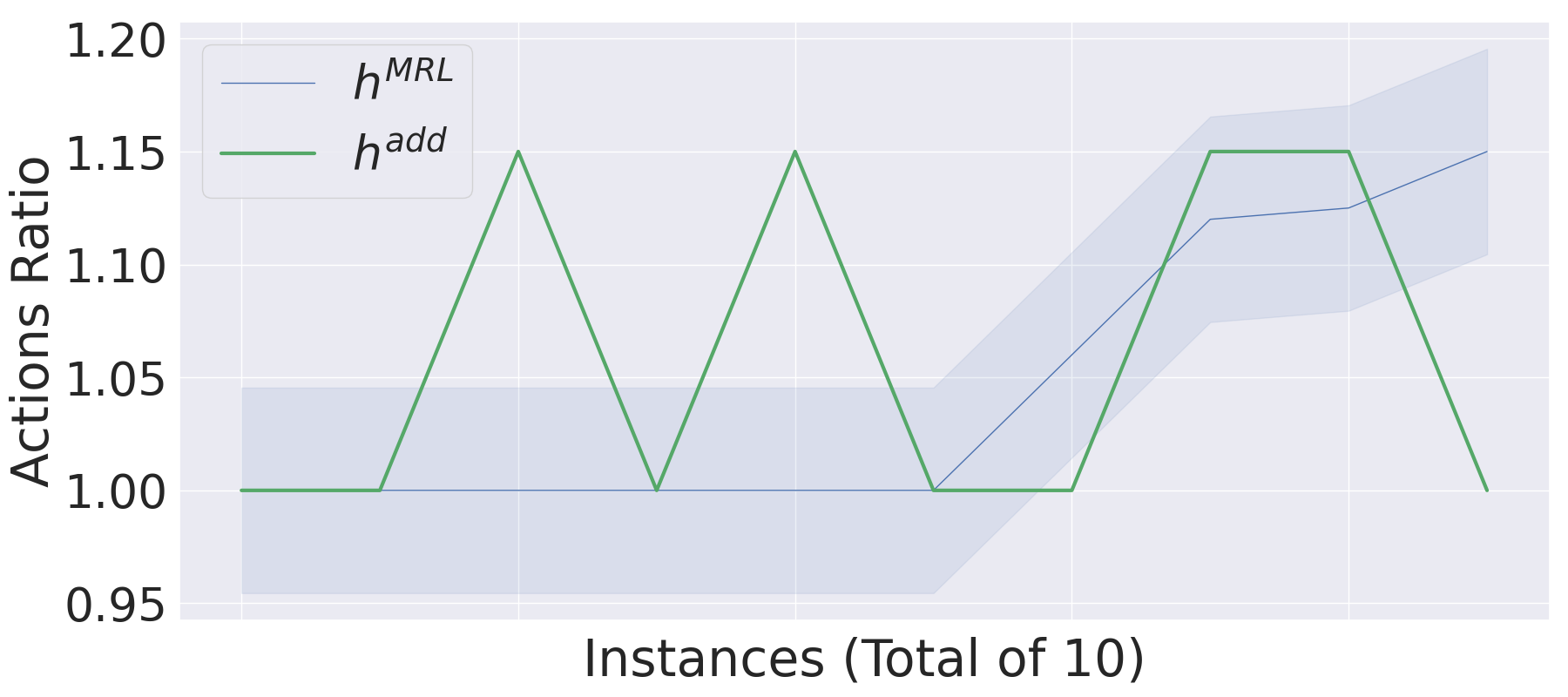}& \\
		
	\end{tabular}
	
	\caption{Comparison of $h^{MRL}$ against domain-independent planning heuristics. Error bars in the bar plot, and shaded areas in the line plot, show the standard deviation of the learning method. Since LM-CUT does not support conditional effects it is not possible to use it in Nurikabe.}
	\label{fig:results} % I can do without the label too
\end{figure*}

\begin{figure*}
	\centering
	\begin{tabular}{ccccc}
		& \hfil \LARGE{Nodes Expanded} & \hfil \LARGE{Task Addition on $h^{SUPER}$} \\
		\raisebox{.9\height}{\rotatebox{90}{\LARGE{Snake}}}&
		\includegraphics[width=8cm, height=3.3cm]{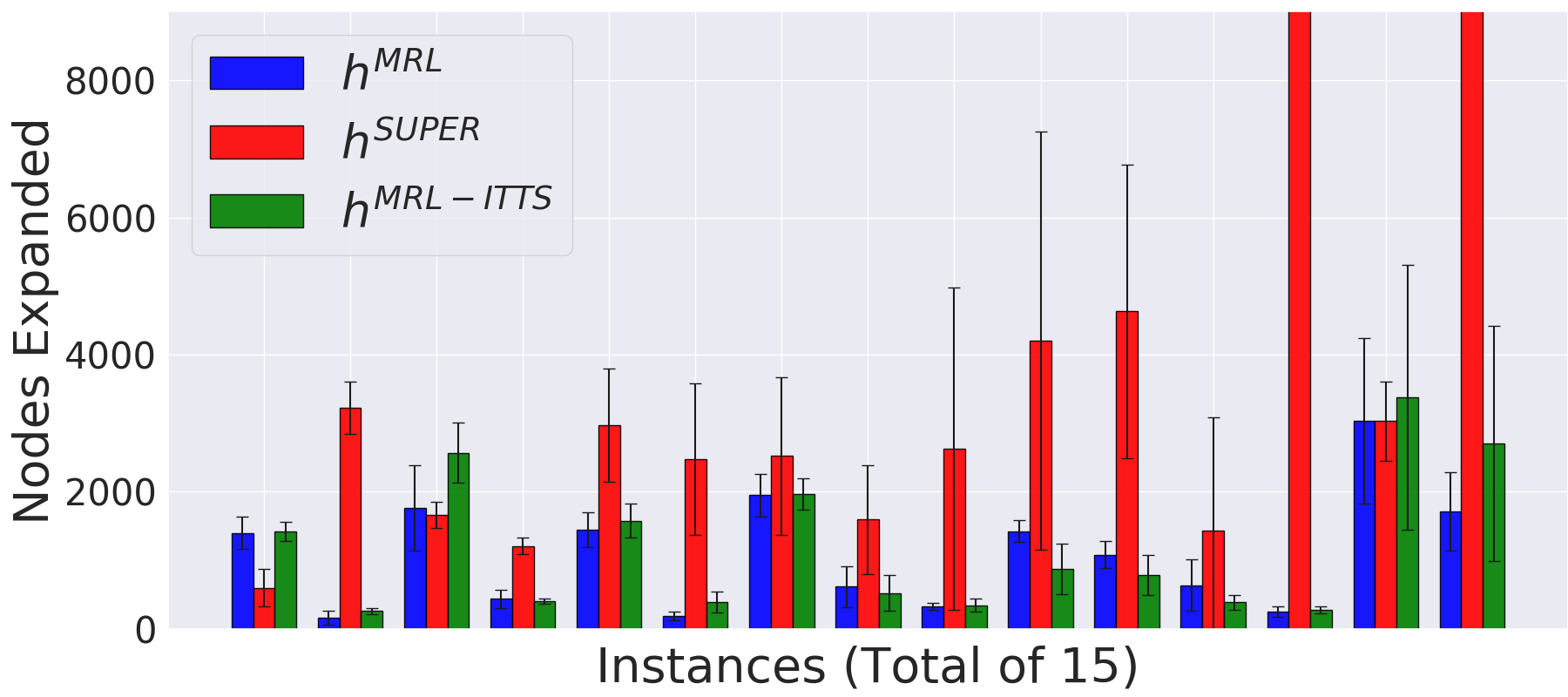}&
		\includegraphics[width=8cm, height=3.3cm]{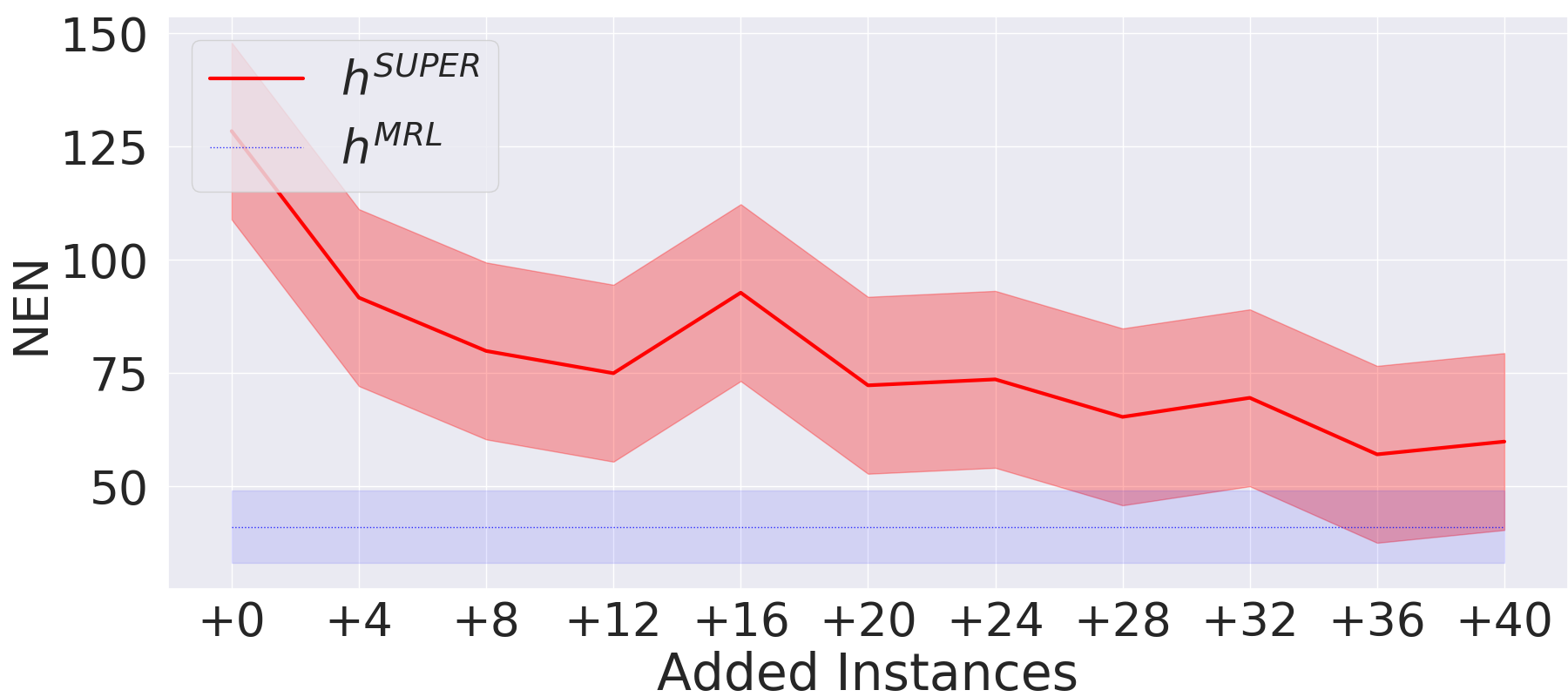}& \\
		\raisebox{.5\height}{\rotatebox{90}{\LARGE{Sokoban}}} &
		\includegraphics[width=8cm, height=3.3cm]{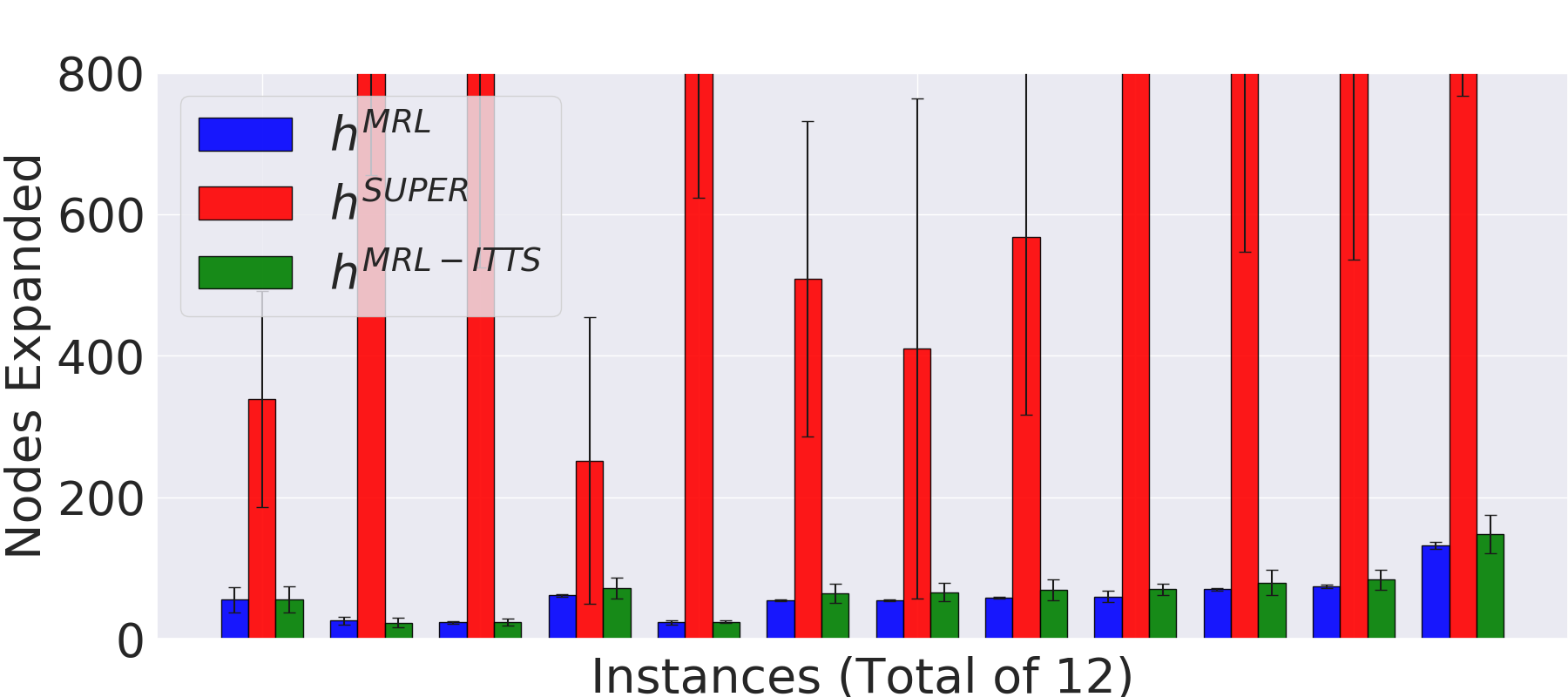}&
		\includegraphics[width=8cm, height=3.3cm]{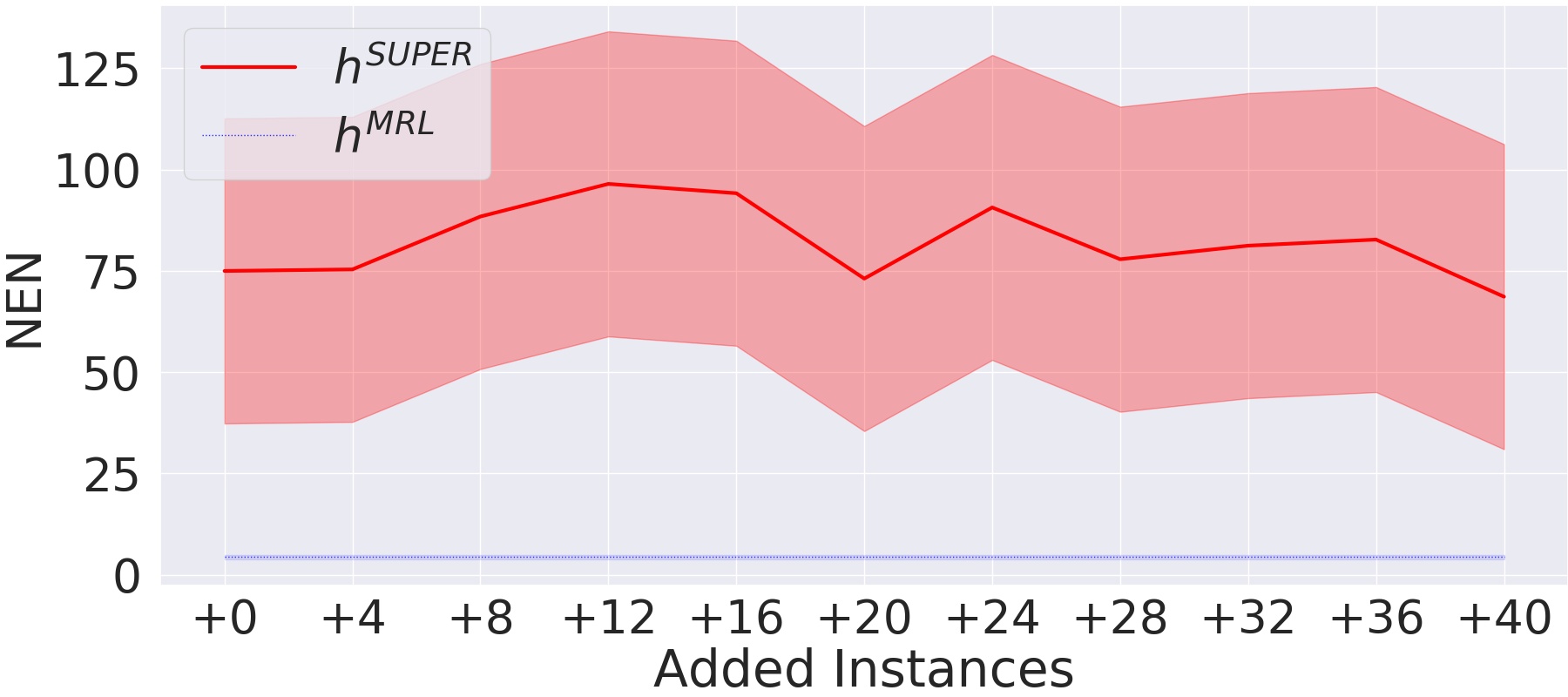}&\\
		\raisebox{.9\height}{\rotatebox{90}{\LARGE{Blocks}}} &
		\includegraphics[width=8cm, height=3.3cm]{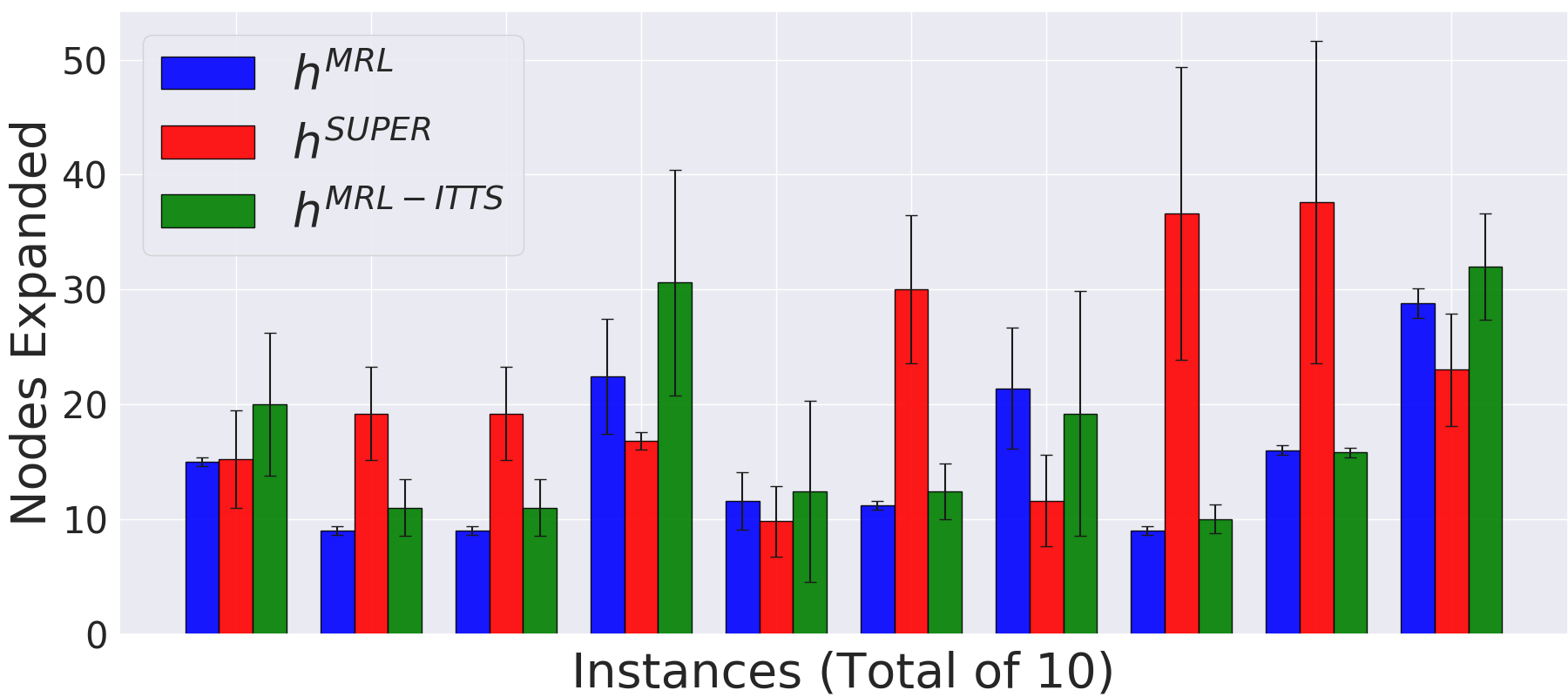}&
		\includegraphics[width=8cm, height=3.3cm]{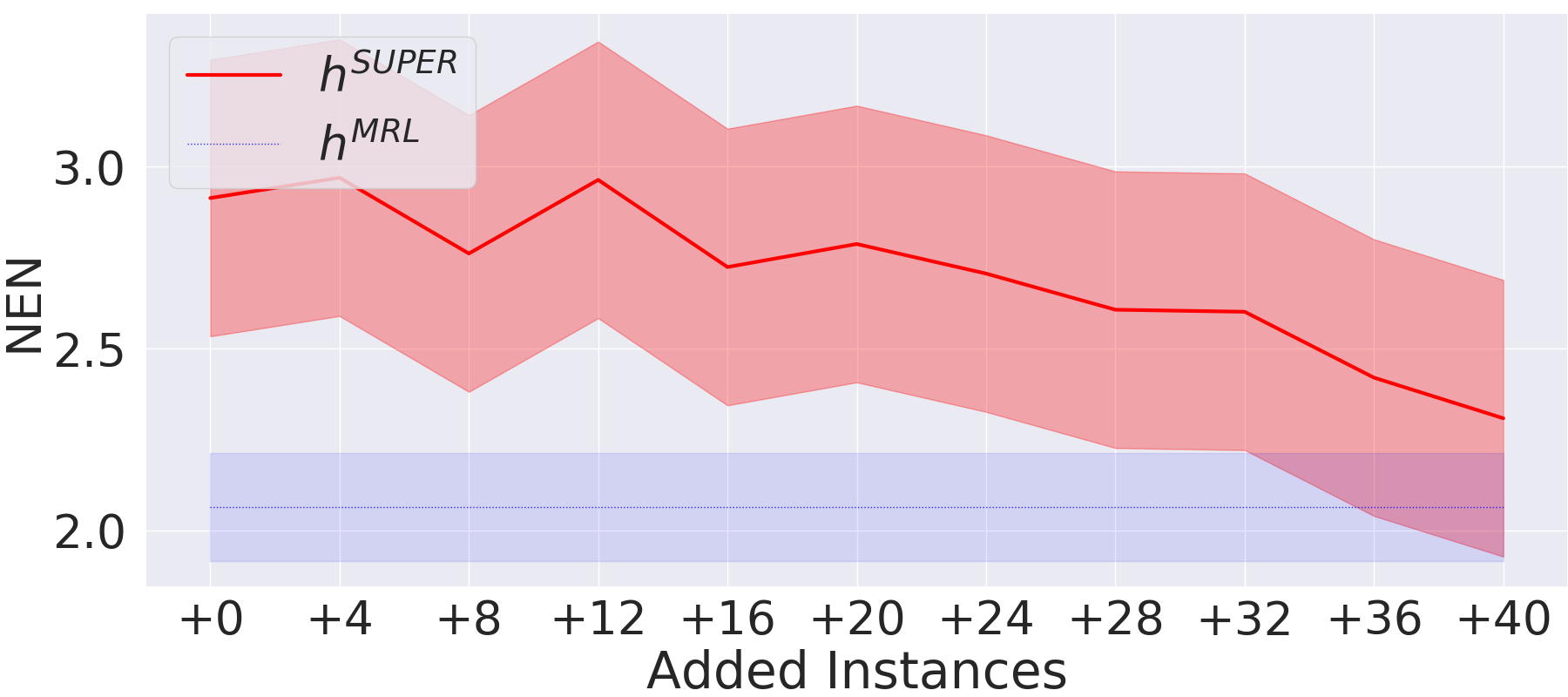}& \\
		\raisebox{.5\height}{\rotatebox{90}{\LARGE{Nurikabe}}} &
		\includegraphics[width=8cm, height=3.3cm]{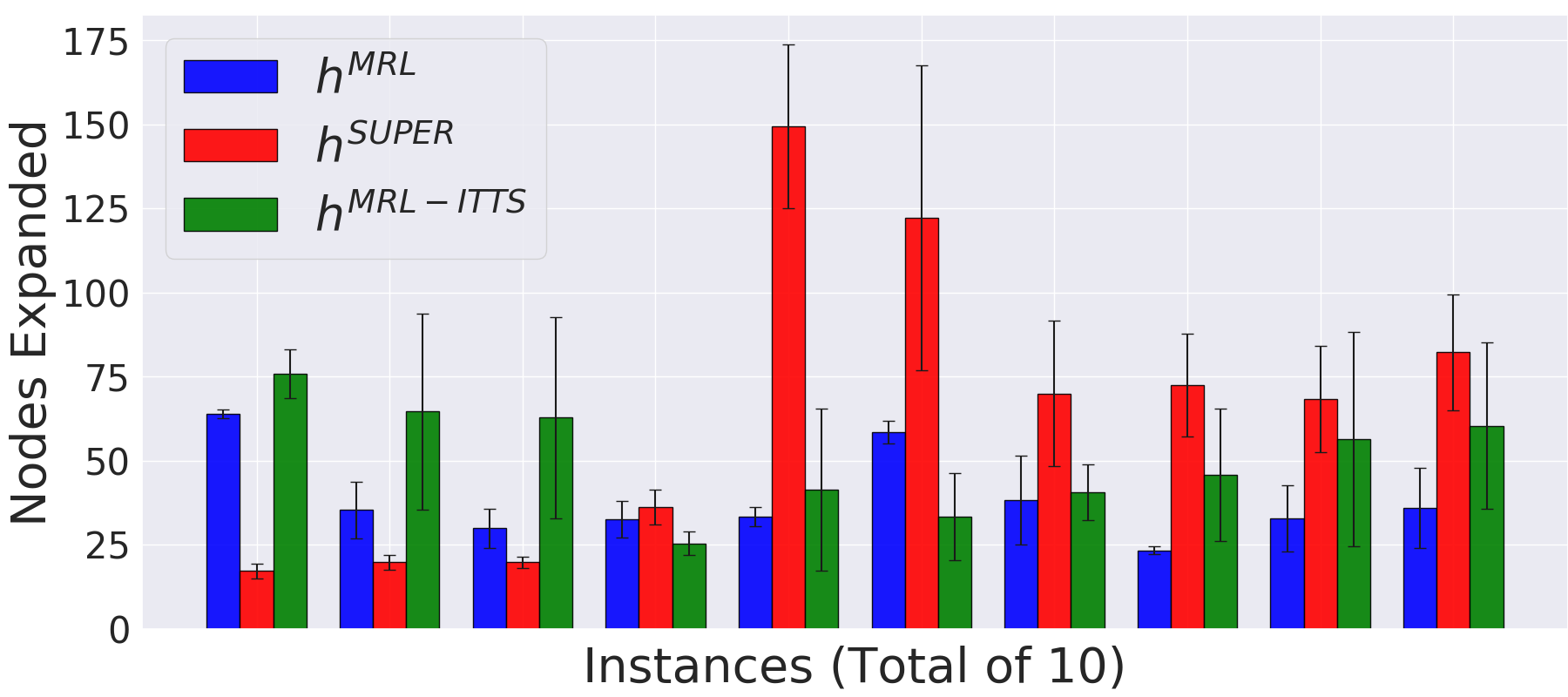}&
		\includegraphics[width=8cm, height=3.3cm]{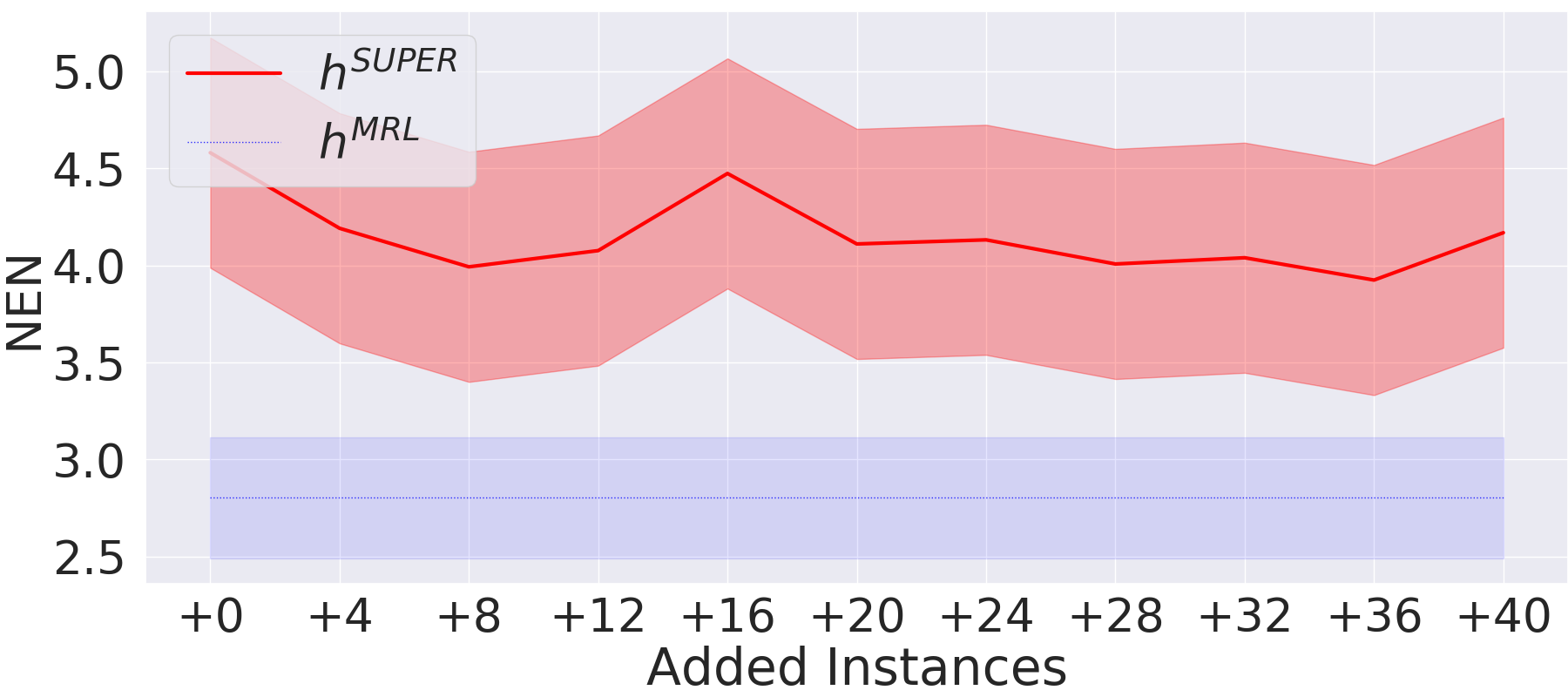}& \\
		\raisebox{1.2\height}{\rotatebox{90}{\LARGE{Ferry}}} &
		\includegraphics[width=8cm, height=3.3cm]{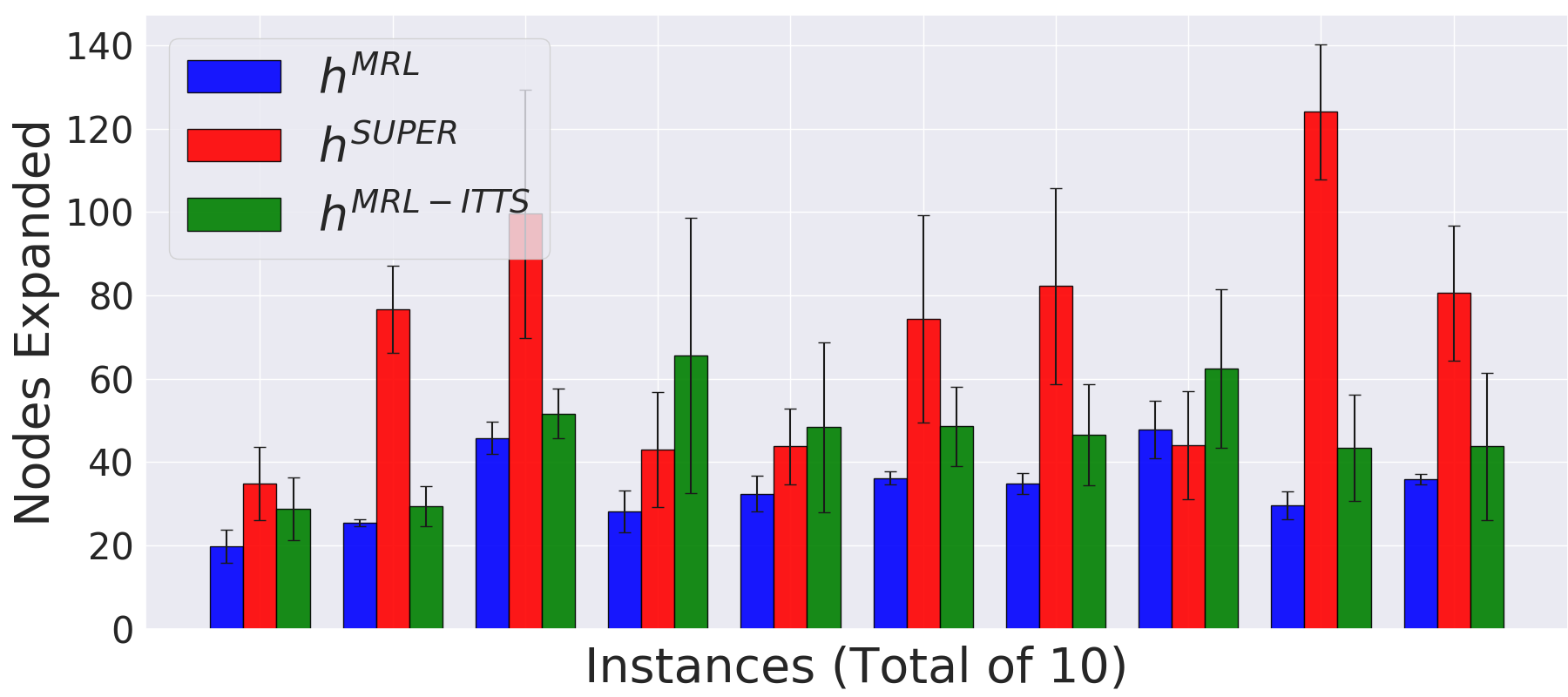}&
		\includegraphics[width=8cm, height=3.3cm]{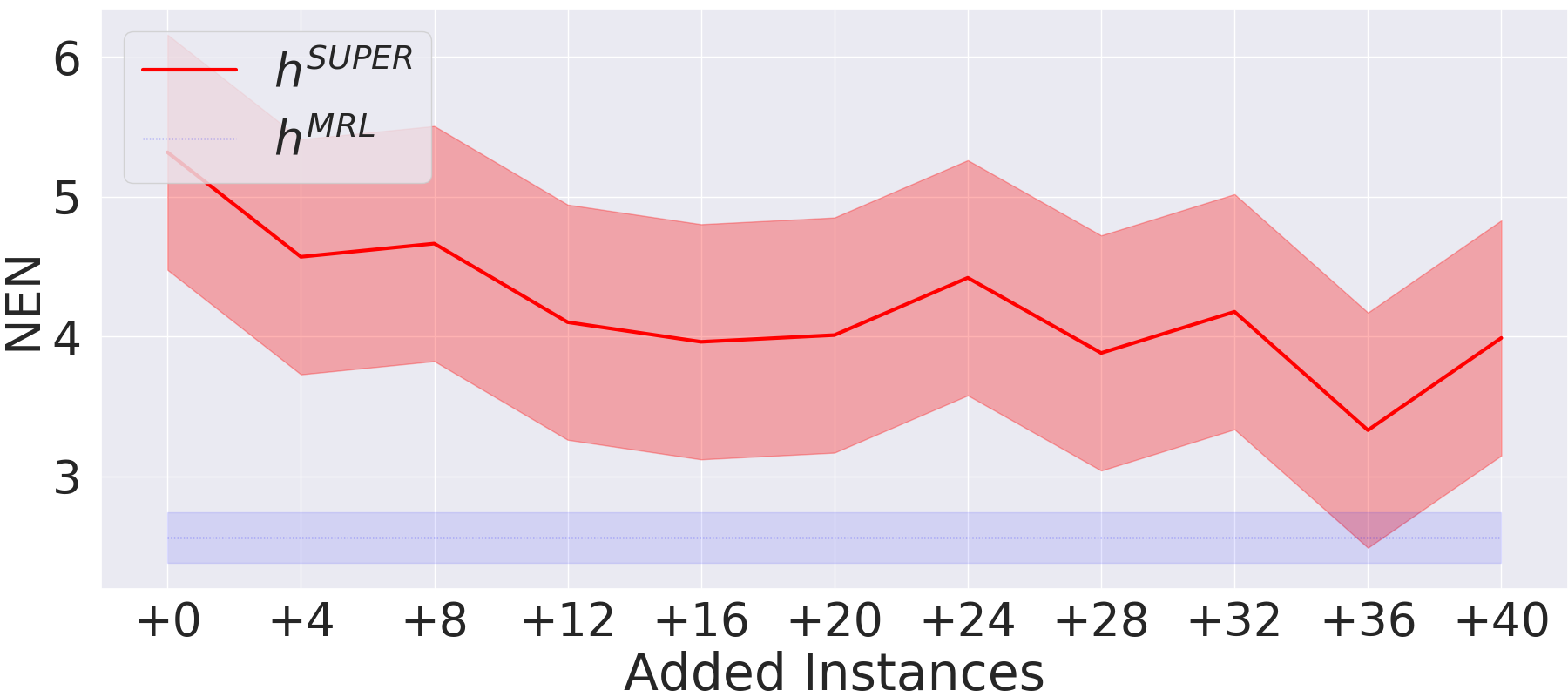}&\\
		\raisebox{.7\height}{\rotatebox{90}{\LARGE{Gripper}}} &
		\includegraphics[width=8cm, height=3.3cm]{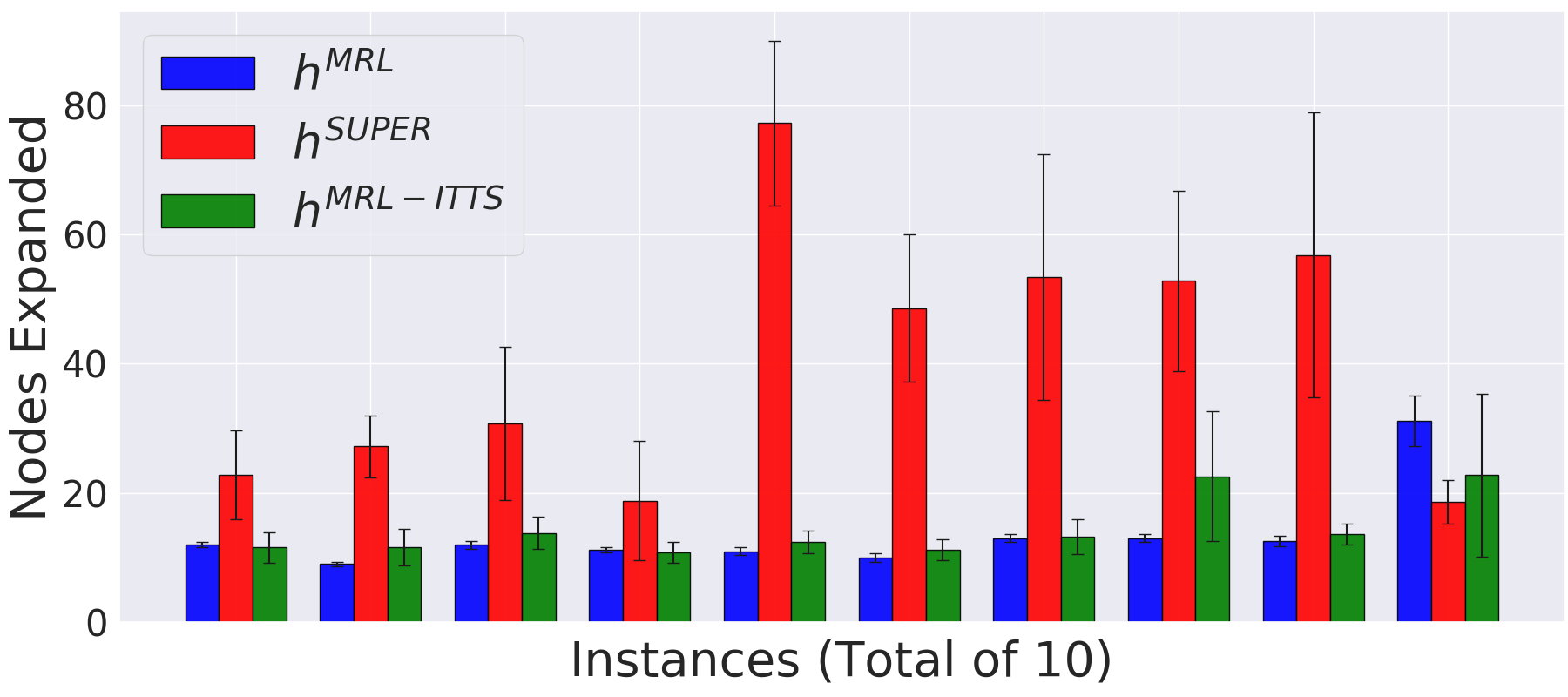}&
		\includegraphics[width=8cm, height=3.3cm]{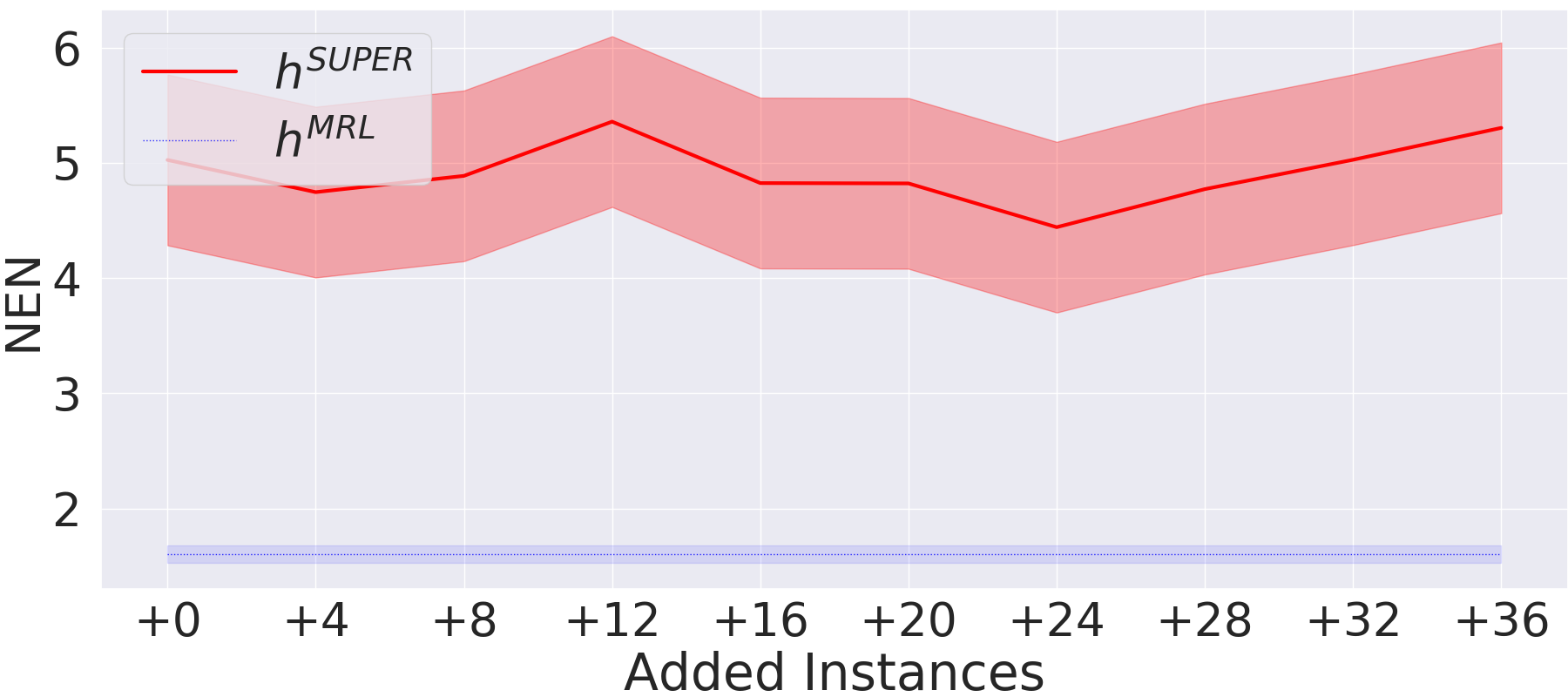}& \\
		
	\end{tabular}
	
	\caption{Comparison of the learning methods.  Error bars in the bar plot represent standard deviation. The shaded area in the Normalized Expanded Nodes (NEN) shows the  95\% confidence interval, since this is the mean over all instances. The number after the $+$ sign refers to the number of tasks added to the original training set for $h^{SUPER}$.}
	\label{fig:results2} 
\end{figure*}

\subsection{Results}

Figure \ref{fig:results} shows the results of the comparison between $h^{MRL}$ and the domain-independent heuristics in all domains. The left-most column of plots shows the nodes expanded by A* for each test instance in each domain, where the blind heuristic gives an indication of the difficulty of the instance. The instances, on the x axis, are sorted in increasing number of states expanded using the blind heuristic. While no heuristic outperforms all the other ones in every instance, $h^{MRL}$ outperforms the baselines in most of them, and on average overall as shown in Table \ref{table:average_performance}. The advantage is particularly evident on hard instances, where the blind heuristic leads to the expansion of the highest number of states. On the other hand, instances that are solved by short plans benefit the least, probably because the internal memory of the function approximator requires a few transitions to generate the context, and stabilize the estimate of the costs. The meta-learned heuristic is therefore less reliable in the first few states. The right-most column of plots shows the length of the computed plans against $h^{add}$, which is also not admissible but quite commonly used, normalized over the length of the optimal plan. The meta-learned heuristic achieves comparable, if not better, plan length overall. Furthermore, all domain-independent heuristics show a performance that is highly dependent on the single instance, while $h^{MRL}$ has quite consistent performance within each domain.

In the left-hand column of Figure \ref{fig:results2}, we show the results on the same instances against the supervised learning approach, and against a meta-learned heuristic without task selection, indicated as $h^{MRL-ITTS}$. The supervised heuristic $h^{SUPER}$ has been trained on the same instances as $h^{MRL}$. ITTS improved the meta-learned heuristic in almost every instance, while at the same time greatly reducing the variance. The meta-learned heuristic outperforms $h^{SUPER}$ in almost every instance, often by a large margin, with the exception of a few instances with short plans, as previously noted. Overall,  $h^{MRL}$ expands less than half the states as $h^{SUPER}$.

The last set of plots, in the right-most column of Figure \ref{fig:results2}, shows the performance of the supervised heuristic as the number of randomly generated training tasks increases. The ``+0'' value in these plots corresponds to the $\mathcal{C}$ training set for each domain, and reported in the domain description above. The y axis reports Normalized Expanded Nodes (NEN), which are the number of expanded states divided by the length of the optimal plan. The normalization allows us to average the results over all instances within the domain. In Snake and Block it takes $40$ additional training tasks for the performance of the supervised heuristic to be comparable. In all other domains even with $40$ additional tasks (which makes a total number of training tasks about $5$ times larger) the supervised heuristic still expands more nodes than $h^{MRL}$ trained over the ``+0'' set.

\begin{table}[]
	
	\centering
	\begin{tabular}{|l|r|}
		
		\hline
		
		\multicolumn{1}{|c|}{\textbf{Heuristic}} & \multicolumn{1}{c|}{\textbf{Expanded States}} \\ \hline
		
		$h^{MRL}$      & 0.0749112                 \\ \hline
		
		$h^{MRL-ITTS}$ & 0.0894698                  \\ \hline
		
		$h^{add}$      & 0.1341158                 \\ \hline
		
		LM-CUT         & 0.1473007                 \\ \hline
		
		$h^{SUPER}$    & 0.1991076                 \\ \hline
		
		$h^{max}$      & 0.3566618                 \\ \hline
		
		Blind          & 1                          \\ \hline
		
	\end{tabular}
	\caption{Average number of expanded states over all domains, normalized with respect to the Blind heuristic.}
	\label{table:average_performance}
\end{table}

\section{Conclusion}

We introduced $h^{MRL}$, a meta-RL heuristic able to generalise in a wide range of planning domains. We show that $h^{MRL}$  outperforms on average both popular domain-independent heuristics, and a supervised learning one. The meta-learning approach appears to be particularly advantageous on hard tasks, while it is less competitive on easier tasks, solved by short plans. A simple solution for this problem would be to expand states according to a domain-independent heuristic while the LSTM memory is filled, and let the meta-RL heuristic take over from there. These results were obtained without any particular encoding of the state space, which instead has received a fair amount of attention in the supervised learning literature. We expect that the results can be further improved with ad-hoc state encodings. Most importantly, this work shows that meta-RL heuristics are viable in planning, and thus creates a new line of heuristic learning.

\bibliography{PhDLiterature}

\begin{thebibliography}{39}
\providecommand{\natexlab}[1]{#1}
\providecommand{\url}[1]{\texttt{#1}}
\providecommand{\urlprefix}{URL }
\expandafter\ifx\csname urlstyle\endcsname\relax
  \providecommand{\doi}[1]{doi:\discretionary{}{}{}#1}\else
  \providecommand{\doi}{doi:\discretionary{}{}{}\begingroup
  \urlstyle{rm}\Url}\fi

\bibitem[{Arfaee, Zilles, and Holte(2011)}]{Arfaee2011}
Arfaee, S.~J.; Zilles, S.; and Holte, R.~C. 2011.
\newblock Learning heuristic functions for large state spaces.
\newblock \emph{Artificial Intelligence} 175: 2075--2098.

\bibitem[{Duan et~al.(2016)Duan, Schulman, Chen, Bartlett, Sutskever, and
  Abbeel}]{Duan2016}
Duan, Y.; Schulman, J.; Chen, X.; Bartlett, P.~L.; Sutskever, I.; and Abbeel,
  P. 2016.
\newblock RL2: Fast reinforcement learning via slow reinforcement learning.
\newblock arXiv preprint arXiv:1611.02779.

\bibitem[{Edelkamp and Schr\"{o}dl(2012)}]{Edelkamp2012}
Edelkamp, S.; and Schr\"{o}dl, S. 2012.
\newblock \emph{Heuristic Search: Theory and Applications}.
\newblock ELSEVIER.

\bibitem[{Fern, Khardon, and Tadepalli(2011)}]{Fern2011}
Fern, A.; Khardon, R.; and Tadepalli, P. 2011.
\newblock The first learning track of the international planning competition.
\newblock \emph{Machine Learning} 84: 81--107.

\bibitem[{Finn, Abbeel, and Levine(2017)}]{Finn2017}
Finn, C.; Abbeel, P.; and Levine, S. 2017.
\newblock Model-agnostic meta-learning for fast adaptation of deep networks.
\newblock In \emph{Proceedings of the 34th International Conference on Machine
  Learning-Volume 70 (ICML)}, 1126--1135.

\bibitem[{Franco et~al.(2018)Franco, Lelis, Barley, Edelkamp, Martinez, and
  Moraru}]{Franco2018}
Franco, S.; Lelis, L. H.~S.; Barley, M.; Edelkamp, S.; Martinez, M.; and
  Moraru, I. 2018.
\newblock The Complementary1 Planner in the IPC 2018.
\newblock In \emph{Association for the Advancement of Artificial Intelligence
  (AAAI)}.

\bibitem[{Garg, Bajpai, and Mausam(2019)}]{Garg2019}
Garg, S.; Bajpai, A.; and Mausam. 2019.
\newblock Size Independent Neural Transfer for RDDL Planning.
\newblock In \emph{International Conference on Automated Planning and
  Scheduling (ICAPS)}.

\bibitem[{Garg, Bajpai, and Mausam(2020)}]{Garg2020}
Garg, S.; Bajpai, A.; and Mausam. 2020.
\newblock Symbolic Network: Generalized Neural Policies for Relational MDPs.
\newblock In \emph{International Conference on Machine Learning (ICML)}.

\bibitem[{Garrett, Kaelbling, and Lozano-P{\'e}rez(2016)}]{Garrett2016}
Garrett, C.~R.; Kaelbling, L.~P.; and Lozano-P{\'e}rez, T. 2016.
\newblock Learning to rank for synthesizing planning heuristics.
\newblock In \emph{International Joint Conferences on Artificial Intelligence
  (IJCAI)}.

\bibitem[{Gomoluch et~al.(2017)Gomoluch, Alrajeh, Russo, and
  Bucchiarone}]{Gomoluch2017}
Gomoluch, P.; Alrajeh, D.; Russo, A.; and Bucchiarone, A. 2017.
\newblock Towards learning domain-independent planning heuristics.
\newblock arXiv preprint arXiv:1707.06895.

\bibitem[{Groshev et~al.(2018)Groshev, Tamar, Goldstein, Srivastava, and
  Abbeel}]{Groshev2018}
Groshev, E.; Tamar, A.; Goldstein, M.; Srivastava, S.; and Abbeel, P. 2018.
\newblock Learning Generalized Reactive Policies Using Deep Neural Networks.
\newblock In \emph{AAAI Conference on Artificial Intelligence}.

\bibitem[{Gupta et~al.(2018)Gupta, Mendonca, Liu, Abbeel, and
  Levine}]{gupta2018meta}
Gupta, A.; Mendonca, R.; Liu, Y.; Abbeel, P.; and Levine, S. 2018.
\newblock Meta-reinforcement learning of structured exploration strategies.
\newblock In \emph{Advances in Neural Information Processing Systems
  (NeurIPS)}, 5302--5311.

\bibitem[{Hart, Nilsson, and Raphael(1968)}]{Hartet1968}
Hart, P.; Nilsson, N.; and Raphael, B. 1968.
\newblock A Formal Basis for the Heuristic Determination of Minimum-Cost Paths.
\newblock \emph{IEEE Transactions of Systems Science and Cyber-netics} .

\bibitem[{Helmert(2006)}]{Helmert2006}
Helmert, M. 2006.
\newblock The Fast Downward Planning System.
\newblock \emph{Journal of Artificial Intelligence Research} .

\bibitem[{Helmert and Domshlak(2010)}]{helmert2009}
Helmert, M.; and Domshlak, C. 2010.
\newblock Landmarks, Critical Paths and Abstractions: What's the Difference
  Anyway?
\newblock In Brim, L.; Edelkamp, S.; Hansen, E.~A.; and Sanders, P., eds.,
  \emph{Graph Search Engineering}, number 09491 in Dagstuhl Seminar
  Proceedings. Dagstuhl, Germany: Schloss Dagstuhl - Leibniz-Zentrum fuer
  Informatik, Germany.
\newblock ISSN 1862-4405.
\newblock \urlprefix\url{http://drops.dagstuhl.de/opus/volltexte/2010/2432}.

\bibitem[{Hoffmann and Nebel(2001)}]{Hoffmann201}
Hoffmann, J.; and Nebel, B. 2001.
\newblock The FF Planning System: Fast Plan Generation Through Heuristic
  Search.
\newblock \emph{Journal of Artificial Intelligence Research} .

\bibitem[{Houthooft et~al.(2018)Houthooft, Chen, Isola, Stadie, Wolski, Ho, and
  Abbeel}]{Houthooft2018}
Houthooft, R.; Chen, R.~Y.; Isola, P.; Stadie, B.~C.; Wolski, F.; Ho, J.; and
  Abbeel, P. 2018.
\newblock Evolved Policy Gradients.
\newblock In \emph{Conference on Neural Information Processing Systems
  (NeurIPS)}.

\bibitem[{Issakkimuthua, Fern, and Tadepalli(2018)}]{Issakkimuthu2018}
Issakkimuthua, M.; Fern, A.; and Tadepalli, P. 2018.
\newblock Training Deep Reactive Policies for Probabilistic Planning Problems.
\newblock In \emph{International Conference on Automated Planning and
  Scheduling (ICAPS)}.

\bibitem[{Katz et~al.(2018)Katz, Sohrabi, Samulowitz, and Sievers}]{Katz2018}
Katz, M.; Sohrabi, S.; Samulowitz, H.; and Sievers, S. 2018.
\newblock Delfi: Online Planner Selection for Cost-Optimal Planning.
\newblock \emph{IPC-9 planner abstracts} 57--64.

\bibitem[{Li et~al.(2018)Li, Yang, Song, and Hospedales}]{Li2018}
Li, D.; Yang, Y.; Song, Y.-Z.; and Hospedales, T.~M. 2018.
\newblock Learning to generalize: Meta-learning for domain generalization.
\newblock In \emph{Thirty-Second AAAI Conference on Artificial Intelligence
  (AAAI)}.

\bibitem[{Luna~Gutierrez and Leonetti(2020)}]{Luna2020}
Luna~Gutierrez, R.; and Leonetti, M. 2020.
\newblock Information-theoretic Task Selection forMeta-Reinforcement Learning.
\newblock In \emph{Conference on Neural Information Processing Systems
  (NeurIPS)}.

\bibitem[{Ma et~al.(2020)Ma, Ferber, Huo, Chen, and Katz}]{Ma2020}
Ma, T.; Ferber, P.; Huo, S.; Chen, J.; and Katz, M. 2020.
\newblock Online Planner Selection with Graph Neural Networks and Adaptive
  Scheduling.
\newblock In \emph{AAAI Conference on Artificial Intelligence (AAAI)}.

\bibitem[{Mishra et~al.(2018)Mishra, Rohaninejad, Chen, and
  Abbeel}]{Mishra2018}
Mishra, N.; Rohaninejad, M.; Chen, X.; and Abbeel, P. 2018.
\newblock A simple neural attentive meta-learner.
\newblock \emph{In Interna-tional Conference on Learning Representations
  (ICLR)} .

\bibitem[{Rakelly et~al.(2019)Rakelly, Zhou, Quillen, Finn, and
  Levine}]{Rakelly2019}
Rakelly, K.; Zhou, A.; Quillen, D.; Finn, C.; and Levine, S. 2019.
\newblock Efficient off-policy meta-reinforcement learning via probabilistic
  context variables.
\newblock \emph{Proceedings of the 36th International Conference on Machine
  Learning (ICML)} .

\bibitem[{Samadi, Felner, and Schaeffer(2008)}]{Samadi2008}
Samadi, M.; Felner, A.; and Schaeffer, J. 2008.
\newblock Bootstrap Learning of Heuristic Functions.
\newblock In \emph{AAAI Conference on Artificial Intelligence (AAAI)}.

\bibitem[{Schulman et~al.(2017)Schulman, Wolski, Dhariwal, Radford, and
  Klimov}]{Schulman2017}
Schulman, J.; Wolski, F.; Dhariwal, P.; Radford, A.; and Klimov, O. 2017.
\newblock Proximal policy optimization algorithms.
\newblock arXiv preprint arXiv:1707.06347.

\bibitem[{Schweighofer and Doya(2003)}]{Schweighofer2003}
Schweighofer, N.; and Doya, K. 2003.
\newblock Meta-learning in reinforcement learning.
\newblock \emph{Neural Networks} 16(1): 5--9.

\bibitem[{Shen, Trevizan, and Thi{\'e}baux(2020)}]{Shen2020}
Shen, W.; Trevizan, F.; and Thi{\'e}baux, S. 2020.
\newblock Learning domain-independent planning heuristics with hypergraph
  networks.
\newblock In \emph{International Conference on Automated Planning and
  Scheduling (ICAPS)}.

\bibitem[{Shen et~al.(2019)Shen, Trevizan, Toyer, Thi{\'e}baux, and
  Xie}]{Shen2019GuidingSW}
Shen, W.; Trevizan, F.~W.; Toyer, S.; Thi{\'e}baux, S.; and Xie, L. 2019.
\newblock Guiding Search with Generalized Policies for Probabilistic Planning.
\newblock In \emph{SOCS}.

\bibitem[{Sievers et~al.(2019)Sievers, Katz, Sohrabi, Samulowitz, and
  Ferber}]{Sievers2019}
Sievers, S.; Katz, M.; Sohrabi, S.; Samulowitz, H.; and Ferber, P. 2019.
\newblock Deep Learning for Cost-Optimal Planning: Task-Dependent Planner
  Selection.
\newblock In \emph{AAAI Conference on Artificial Intelligence (AAAI)}.

\bibitem[{Stadie et~al.(2018)Stadie, Yang, Houthooft, Chen, Duan, Wu, Abbeel,
  and Sutskever}]{Stadie2018}
Stadie, B.; Yang, G.; Houthooft, R.; Chen, P.; Duan, Y.; Wu, Y.; Abbeel, P.;
  and Sutskever, I. 2018.
\newblock The importance of sampling inmeta-reinforcement learning.
\newblock In \emph{Advances in Neural Information Processing Systems
  (NeurIPS)}.

\bibitem[{Sung et~al.(2017)Sung, Zhang, Xiang, and Hospedales}]{Sung2017}
Sung, F.; Zhang, L.; Xiang, T.; and Hospedales, T. 2017.
\newblock Learning to Learn: Meta-Critic Networks for Sample Efficient
  Learning.
\newblock arXiv preprint arXiv:1706.09529v1.

\bibitem[{Thayer, Dionne, and Ruml(2011)}]{Thayer2011}
Thayer, J.~T.; Dionne, A.; and Ruml, W. 2011.
\newblock Learning Inadmissible Heuristic During Search.
\newblock In \emph{International Conference on Automated Planning and
  Scheduling (ICAPS)}.

\bibitem[{Toyer et~al.(2018)Toyer, Trevizan, Thi{\'e}baux, and Xie}]{Toyer2018}
Toyer, S.; Trevizan, F.; Thi{\'e}baux, S.; and Xie, L. 2018.
\newblock Action schema networks: Generalised policies with deep learning.
\newblock In \emph{AAAI Conference on Artificial Intelligence}.

\bibitem[{Vanschoren(2018)}]{Vanschoren2018}
Vanschoren, J. 2018.
\newblock Meta-Learning: A Survey.
\newblock arXiv preprint arXiv:1810.03548.

\bibitem[{Wang et~al.(2017)Wang, Kurth-Nelson, Tirumala, Soyer, Leibo, Munos,
  Blundell, Kumaran, and Botivnick}]{Wang2017}
Wang, J.; Kurth-Nelson, Z.; Tirumala, D.; Soyer, H.; Leibo, J.; Munos, R.;
  Blundell, C.; Kumaran, D.; and Botivnick, M. 2017.
\newblock Learning to reinforcement learn.
\newblock arXiv preprint arXiv:1611.05763.

\bibitem[{Xu et~al.(2018)Xu, Liu, Zhao, and Peng}]{Xu2018}
Xu, T.; Liu, Q.; Zhao, L.; and Peng, J. 2018.
\newblock Learning to Explore via Meta-Policy Gradient.
\newblock In \emph{International Conference on MachineLearning (ICML)}.

\bibitem[{Xu, van Hassel, and Silver(2018)}]{Xu22018}
Xu, Z.; van Hassel, H.; and Silver, D. 2018.
\newblock Meta-Gradient Reinforcement Learning.
\newblock In \emph{Conference on Neural Information Processing System
  (NeurIPS)}.

\bibitem[{Yoon, Fern, and Givan(2008)}]{Yoon2008}
Yoon, S.; Fern, A.; and Givan, R. 2008.
\newblock Learning control knowledge for forward search planning.
\newblock \emph{Journal of Machine Learning Research} 9: 683--718.

\end{thebibliography}

\end{document}